\documentclass[journal]{IEEEtran}
\ifCLASSINFOpdf
  \usepackage[pdftex]{graphicx}
\else
\fi
\usepackage{amssymb}
\usepackage[cmex10]{amsmath}
\newtheorem{thm}{Theorem}
\newtheorem{defn}{Definition}

\newtheorem{lema}{Lemma}

\newtheorem{remk}{Remark}

\usepackage{algorithm}
\usepackage{algorithmic}

\usepackage{cases}
\usepackage{array}

\usepackage{booktabs}

\usepackage[tight,footnotesize]{subfigure}

\usepackage{amssymb}
\usepackage{cases}

\begin{document}
%
\title{Multi-View Matrix Completion for Multi-Label Image Classification}
%
%
%

\author{Yong~Luo,
        Tongliang~Liu,
        Dacheng~Tao,~\IEEEmembership{Senior Member,~IEEE,}
        and~Chao~Xu,~\IEEEmembership{Member,~IEEE}
\thanks{Y. Luo and C. Xu are with the Key Laboratory of Machine Perception (Ministry of Education),
School of Electronics Engineering and Computer Science, Peking University, Beijing, 100871, China.}
\thanks{T. Liu and D. Tao are affiliated with the Centre for Quantum Computation \& Intelligent Systems and the Faculty of Engineering \& Information Technology, University of Technology, Sydney, 235 Jones Street, Ultimo, NSW 2007, Australia.}
\thanks{\copyright 2015 IEEE. Personal use of this material is permitted. Permission from IEEE must be obtained for all other uses, in any current or future media, including reprinting/republishing this material for advertising or promotional purposes, creating new collective works, for resale or redistribution to servers or lists, or reuse of any copyrighted component of this work in other works.}
}

%
%



\markboth{$>$ \normalsize{TIP-12151-2014 R}\footnotesize{evision} \normalsize{1} $<$}%
{Shell \MakeLowercase{\textit{et al.}}: Bare Demo of IEEEtran.cls for Journals}

%



\maketitle

\begin{abstract}
There is growing interest in multi-label image classification due to its critical role in web-based image analytics-based applications, such as large-scale image retrieval and browsing. Matrix completion has recently been introduced as a method for transductive (semi-supervised) multi-label classification, and has several distinct advantages, including robustness to missing data and background noise in both feature and label space. However, it is limited by only considering data represented by a single-view feature, which cannot precisely characterize images containing several semantic concepts. To utilize multiple features taken from different views, we have to concatenate the different features as a long vector. But this concatenation is prone to over-fitting and often leads to very high time complexity in MC based image classification. Therefore, we propose to weightedly combine the MC outputs of different views, and present the multi-view matrix completion (MVMC) framework for transductive multi-label image classification. To learn the view combination weights effectively, we apply a cross validation strategy on the labeled set. In particular, MVMC splits the labeled set into two parts, and predicts the labels of one part using the known labels of the other part. The predicted labels are then used to learn the view combination coefficients. In the learning process, we adopt the average precision (AP) loss, which is particular suitable for multi-label image classification, since the ranking based criteria are critical for evaluating a multi-label classification system. A least squares loss formulation is also presented for the sake of efficiency, and the robustness of the algorithm based on the AP loss compared with the other losses is investigated. Experimental evaluation on two real world datasets (PASCAL VOC' 07 and MIR Flickr) demonstrate the effectiveness of MVMC for transductive (semi-supervised) multi-label image classification, and show that MVMC can exploit complementary properties of different features and output-consistent labels for improved multi-label image classification.
\end{abstract}


%
\IEEEpeerreviewmaketitle

\section{Introduction}
\label{sec:Introduction}
%
%
%
%

\IEEEPARstart{M}{}ulti-label image classification, where multiple labels are assigned to a given image, is useful in many web-based image analytic-based applications. For example, keywords can be automatically assigned to an uploaded web image so that annotated images may be searched directly using text-based image retrieval systems.

Dozens of multi-label algorithms have been proposed in the past decade~\cite{M-Boutell-et-al-PR-2004, G-Tsoumakas-and-I-Katakis-IJDWM-2007, B-Hariharan-et-al-ICML-2010, L-Sun-et-al-TPAMI-2011, B-Wang-et-al-ICCV-2013, G-Sundaramoorthi-and-BW-Hong-CVPR-2014}. However, none of these methods are able to handle missing features, or cases where parts of the training data labels are unknown. Many of the algorithms lack robustness to outliers and background noise. In order to overcome these limitations, matrix completion (MC) has recently been introduced as an alternative methodology for transductive (semi-supervised) multi-label classification~\cite{A-Goldberg-et-al-NIPS-2010, R-Cabral-et-al-NIPS-2011}. In particular, the MC-based multi-label classification concatenates the feature and label matrices, and then completes the unknown entries (either features or labels) in the concatenated matrix by the use of the rank minimization criterion. In this way, the MC-based methods can be used not only to infer labels of the unlabeled data, but also to estimate values of the missing features, and denoise the observed features and labels.

Although MC-based algorithms are robust for general transductive multi-label classification tasks, they cannot directly handle those image classification problems that include images represented by multi-view features. A popular solution has been to concatenate all the features into a long vector, but this strategy not only ignores the physical interpretations of different features, but also encounters an over-fitting problem given frequently limited labeled training samples and high dimensional image features. Besides, the feature concatenation often leads to a very large matrix to be completed. Thus the time cost is very high and sometimes intolerable \cite{J-Cai-et-al-SIAM-2010}.

To avoid these drawbacks, we propose to weightedly combine the MC based classification outputs of different views, and develop a new framework, namely, multi-view matrix completion (MVMC) for handling multi-view features in semi-supervised multi-label image classification. To learn the view combination coefficients, we firstly perform a two-fold cross validation procedure on the labeled set for each view. That is, we divide the labeled training samples into two (usually equal) sets. Labels in one set are assumed to be unknown and are completed using the label information in the other set. In fact the labeled training samples have been annotated, and thus we propose to linearly combine the predicted labels of all the views to approximate the ground-truth labels. In this way, we learn the combination coefficients of different views. In the learning of the coefficients, we propose to directly optimize average precision (AP), which is a critical criterion in evaluating a multi-label classification algorithm. We also present a formulation that adopts the least squares (LS) loss, which is quite efficient although not so proper as the AP loss for multi-label classification. Finally, for each view, the labels of the unlabeled and test data are predicted using matrix completion, and the obtained predictions of all the views are combined using the learned coefficients. The proposed algorithms tend to assign higher weight to the view carrying more discriminative information, and thus explore the complementary nature of different views.

In order to evaluate our MVMC algorithms (MVMC-LS and MVMC-AP), we have used two challenging datasets, PASCAL VOC' 07 \cite{Pascal-VOC-2007} and MIR Flickr \cite{MIR-Flickr-2008}. To the best of our knowledge, there are no other algorithms which employ multi-view matrix completion. Therefore, in order to assess performance, we first compared our algorithms with the best single view (BMC), concatenation of all the views (CMC), and average the outputs of different views (AMC) in terms of mean average precision (mAP), mean area under the ROC curve (mAUC) and hamming loss (HL). To further verify the effectiveness of MVMC, we compared MVMC-AP with some popular and competitive feature-level \cite{A-Rakotomamonjy-et-al-JMLR-2008, M-Kloft-et-al-JMLR-2011} and classifier-level \cite{J-Kludas-et-al-AMR-2008} multi-view approaches, as well as some competitive multi-label classification methods \cite{L-Sun-et-al-TPAMI-2011, B-Wang-et-al-ICCV-2013}. The experimental results show that our novel approach outperforms the current state-of-the-art.

The main contributions of this paper are: 1) the cross validation strategy that learns the view combination coefficients in the proposed multi-view matrix completion framework for multi-label image classification; 2) the developed solutions for the constrained optimization problems with the AP loss, as well as the LS loss. The former is particular suitable for multi-label classification, and the latter is for the sake of efficiency; 3) the robustness analysis of the AP loss compared with the LS and hinge loss in MVMC.

The rest of the paper is organized as follows. We firstly review some related work on matrix completion and multi-view learning in Section \ref{sec:Related_Work}. Section \ref{sec:Transduction_with_MC} summarizes the recent work on utilizing matrix completion for transduction and multi-label classification. In Section \ref{sec:Multiview_MC}, we present the proposed MVMC framework, as well as the MVMC-LS and MVMC-AP algorithms by choosing different loss functions. Moreover, the robustness of the different algorithms is analyzed. The experimental results are presented in Section \ref{sec:Experiments}. Finally, we conclude this paper in Section \ref{sec:Conclusion} and prove the main theorem of this paper in Section \ref{sec:Proofs}.


\section{Related Work}
\label{sec:Related_Work}

\subsection{Matrix completion}

Matrix completion (MC), as the name suggested, is to fill in the unknown entries of an uncompleted matrix $M$. Without any assumption about the nature of the matrix, the completion is impossible. We usually assume the matrix we want to recover is of low-rank~\cite{S-Ono-et-al-TIP-2014}. Then our goal is to find a matrix $X$ so that the errors between $X$ and $M$ on the known entries are as small as possible, and the rank of $X$ is minimized. The rank minimization~\cite{H-Arguello-and-GR-Arce-TIP-2013} problem is NP-hard and thus is of little practical use~\cite{E-Candes-and-B-Recht-FCM-2009}. Fortunately, $\mathrm{rank}(X)$ can be replaced by its convex envelope, the nuclear norm $\|X\|_*$, which is the sum of $X$'s singular values~\cite{M-Fazel-Thesis-2002}. On the basis of this relaxation, dozens of methods have been proposed for matrix completion~\cite{E-Candes-and-B-Recht-FCM-2009, J-Cai-et-al-SIAM-2010, SQ-Ma-et-al-MP-2011, R-Keshavan-et-al-TIT-2010}. A representative work was done by Candes and Recht~\cite{E-Candes-and-B-Recht-FCM-2009}. They showed that the minimization of $\|X\|_*$ and $\mathrm{rank}(X)$ has the same solution under broad assumptions of incoherence, and proved that a limited number of samples are needed to recover a low-rank matrix. Besides, a semi-definite programming (SDP) algorithm was proposed to efficiently solve the nuclear norm minimization problem. However, this algorithm cannot handle large size matrix, and thus a singular value thresholding (SVT) algorithm was developed in~\cite{J-Cai-et-al-SIAM-2010}. SVT can be expressed exactly as a linearized Bregman iteration, and the key step is the derived matrix shrinkage operation. Although SVT is shown to be efficient when completing large size matrices, it may fail (diverges, or does not solve the problems in a tolerate time, or yields a very inaccurate solution) for matrices which are not of very low-rank (e.g., an $1000 \times 1000$ matrix of rank $50$). Therefore, a more robust algorithm, fixed point continuation (FPC)~\cite{SQ-Ma-et-al-MP-2011}, was proposed. Similar to SVT, the matrix shrinkage is also utilized in FPC. The difference is that FPC employed an operator splitting technique. To reduce the complexity and improve recoverability (the ability to recover matrices of either small or moderate rank with tolerate error and speed), the authors further presented an approximate SVD based FPC (FPCA), which was demonstrated empirically to have much better recoverability than SDP (for MC) and SVT, etc. For example, the authors show that ``for matrices of size $1000 \times 1000$ and rank $50$, FPCA can recover them with a relative error of $10^{-5}$, in about $3$ minutes by sampling only $20$ percent of the matrix elements''. But SDP and SVT, etc. do not have such a good recoverability property (since in these methods, the algorithm diverge, or the time cost is intolerant, or the solution is inaccurate).

To recovery a low-rank matrix corrupted with arbitrary large errors, a combination of the nuclear norm and the $l_1$-norm should be minimized. Lin et al. \cite{ZC-Lin-et-al-NIPS-2011} extended the classical augmented Lagrange multipliers (ALM) for solving this minimization problem efficiently. In particular, the exact ALM (EALM) method proposed in \cite{ZC-Lin-et-al-NIPS-2011} was proved to have a pleasing convergence speed. The improved version, inexact ALM (IALM), was shown to be more precise and much faster than the state-of-the-art solvers, such as the accelerated proximal gradient (APG) algorithm \cite{ZC-Lin-et-al-TR-UILU-2009}. Recently, matrix completion was introduced for transductive (semi-supervised) multi-label learning and we will depict it in section \ref{sec:Transduction_with_MC}.

\subsection{Multi-view learning}

Multi-view learning is an active research topic in recent years. The multiple views can be the different viewpoints of an object in the camera, or the various descriptions of a given sample. We focus on the latter in this paper, and the goal is to learn to fuse the different descriptions. Lots of methods have been proposed in the recent decades for multi-view classification~\cite{A-Zien-and-C-Ong-ICML-2007}, retrieval~\cite{J-Kludas-et-al-AMR-2008}, clustering~\cite{S-Bickel-and-T-Scheffer-ICDM-2004}, etc. In this section, we mainly review the classification methods~\cite{A-Rodriguez-et-al-TIP-2013, J-Lindblad-and-N-Sladoje-TIP-2014, R-Ptucha-and-AE-Savakis-TIP-2014}, although most of them are also amenable for other applications. According to the level of the fusion being carried out, the multi-view classification methods can be grouped into two major categories: feature-level fusion and classifier-level fusion. We further divide them into four sub-categories: similarity fusion and unified subspace learning for the feature level, output/decision fusion and interactive fusion for the classifier level.

\subsubsection{Feature-level fusion}
A direct strategy for feature-level fusion is to concatenate the different kinds of features into a long vector. This often leads to the curse of dimensionality problem and thus it is not practical. To this end, many sophisticated techniques are developed, which include those similarity space fusion (mostly kernel fusion by now) and multi-view subspace learning approaches.

\textbf{Similarity space fusion:} As far as we know, most of the current works on similarity space fusion are implemented in the form of kernel fusion. Multiple kernel learning (MKL)~\cite{G-Lanckriet-et-al-ICML-2002} is one of the most representative framework for kernel fusion. For example, in~\cite{G-Lanckriet-et-al-JMLR-2004}, a combination of different kernels built on different features sets were utilized for protein prediction. MKL was also used for dimensionality reduction of the multi-view data based on graph embedding~\cite{Y-Lin-et-al-NIPS-2008}. McFee and Lanckriet~\cite{B-McFee-and-G-Lanckriet-JMLR-2011} presented a method to combine multiple kernels in a proposed partial order embedding algorithm. The method learns a set of kernel mappings to induce a unified multi-modal similarity space, where the human perceptual information expressed by relative comparisons is incorporated. Kloft et al.~\cite{M-Kloft-et-al-JMLR-2011} extended the traditional $l_1$-norm MKL to arbitrary norms, and showed that the non-sparse MKL was superior to the state-of-the-art in combining different feature sets for biometrics recognition.

\textbf{Unified subspace learning:} In the feature-level fusion, another set of algorithms is on multi-view subspace learning. Canonical correlation analysis (CCA)~\cite{D-Hardoon-et-al-NCn-2004} is one of the most popular methods for two-view learning, and seeks a subspace where the given two views are maximally correlated. SVM-2K~\cite{J-Farquhar-et-al-NIPS-2005} combines Kernel CCA (KCCA) and support vector machine (SVM) in a single optimization problem. Recently, White et al.~\cite{M-White-et-al-NIPS-2012} proposed a convex formulation for learning a shared subspace of multiple sources. In the learned subspace, conditional independence constraints are enforced.

\subsubsection{Classifier-level fusion}
Schemes in this category either learn classifiers of different views independently or interactively.

\textbf{Output or decision fusion:} Individual classifiers are created for different views and then the outputs or decisions are fused. In~\cite{C-Snoek-et-al-MM-2005}, the SVM outputs are firstly converted to probabilistic scores, and then concatenated as the input of an SVM for final classification. This method was shown to outperform the simple feature concatenation, followed by an SVM. Such an approach was called hierarchical SVM in~\cite{J-Kludas-et-al-AMR-2008}, and is compared with several other popular classifier fusion strategies, such as weighted sum of outputs and majority voting. Fumera and Roli~\cite{G-Fumera-and-F-Roli-TPAMI-2005} gave a theoretical analysis of the linear combination of multiple classifiers, and the effectiveness of their analytical model was confirmed by the experimental results. A thorough study on the weighted voting methods for classifier fusion was presented in~\cite{M-Wozniak-and-K-Jackowski-HAIS-2009}, where the neural network was adopted to estimate the combination weights.

\textbf{Interactive fusion:} Methods in this sub-category communicate information with other views when learning classifier of the current view. Lots of these methods are semi-supervised and naturally two-view, such as co-training~\cite{A-Blum-and-T-Mitchell-COLT-1998}, co-regularization~\cite{B-Krishnapuram-et-al-NIPS-2004}, etc. In the co-training framework, unlabeled samples classified by one view with high confidence were put in the labeled pool of the other view. Such a process was repeated until the classification performance on the validation dataset decreased. Although succeed empirically, co-training is based on the compatibility and class conditional independence assumptions, which are usually too restrictive to be satisfied~\cite{C-Christoudias-et-al-CVPR-2009}. The co-EM algorithm~\cite{K-Nigam-and-R-Ghani-CIKM-2000} bootstrapped samples in a similar way like co-training, but the unlabeled samples were labeled probabilistically in a batch mode using expectation-maximization (EM). By formulating the linear classifier in a probabilistic framework, SVM was introduced as the base classifier in co-EM~\cite{U-Brefeld-and-T-Scheffer-ICML-2004}. In~\cite{B-Krishnapuram-et-al-NIPS-2004}, classifiers of different views are enforced to be agreed on unlabeled data by the use of a regularization term, in a graph-based semi-supervised framework. This is called co-regularization and a similar idea was utilized in~\cite{V-Sindhwani-et-al-ICMLw-2005}, under the theme of manifold regularization~\cite{M-Belkin-et-al-JMLR-2006}.

It was shown in \cite{C-Snoek-et-al-MM-2005, J-Kludas-et-al-AMR-2008} that classifier-level fusion outperforms simple feature concatenation, while sophisticated feature-level fusion can usually be better than classifier-level fusion since the raw information is preserved \cite{A-Klausner-et-al-ICDSC-2007}. The proposed MVMC method belongs to the classifier-level fusion, and is particular suitable for transductive (semi-supervised) multi-label classification. It was demonstrated empirically in this paper that MVMC can outperform some competitive classifier-level and feature-level multi-view learning approaches when limited labeled data are available. Experiments also show that MVMC is superior to competitive multi-label \cite{L-Sun-et-al-TPAMI-2011} and the recently proposed semi-supervised multi-label classification method \cite{B-Wang-et-al-ICCV-2013}.

\section{Transduction with matrix completion}
\label{sec:Transduction_with_MC}

This section briefly introduces the framework of utilizing matrix completion for transductive (semi-supervised) multi-label learning~\cite{A-Goldberg-et-al-NIPS-2010}. Given a set of $n$ samples $D_n = {(\mathrm{x}_j^0, \mathrm{y}_j^0)_{j=1}^n}$ with each $\mathrm{x}_j^0 \in \mathbb{R}^d$, and the corresponding label vector $\mathrm{y}_j^0 \in \{-1,1\}^m$. We construct the feature matrix $X^0 = [\mathrm{x}_1^0, \ldots, \mathrm{x}_n^0] \in \mathbb{R}^{d \times n}$ and the label matrix $Y^0 = [\mathrm{y}_1^0, \ldots, \mathrm{y}_n^0] \in \mathbb{R}^{m \times n}$. Some entries of $X^0$ may be missing, and our aim is to predict the unknown entries in $Y^0$. To solve this problem, we consider the linear classification model
\begin{equation}
\label{eq:Linear_Cls_Model}
  \mathrm{y}_j = W \mathrm{x}_j+\mathrm{b} = [W\ \mathrm{b}]\left[\begin{array}{c} \mathrm{x}_j \\ 1 \end{array} \right],
\end{equation}
where $W \in \mathbb{R}^{m \times d}$ is the weight matrix, $\mathrm{b} \in \mathbb{R}^m$ is the bias vector, $\mathrm{y}_j \in \mathbb{R}^m$ contains the soft labels of the $j$'th sample and the hard label $y_{ij}^0$ can be obtained by $\mathrm{sign}(y_{ij})$. Then we can construct the underlying feature matrix $X = [\mathrm{x}_1, \ldots ,\mathrm{x}_n]$, the soft label matrix $Y = [y_1, \ldots ,y_n]$, and concatenate them into an $(m+d+1) \times n$ matrix $Z = [Y; X; \mathrm{1}^T]$. The stacked matrix is of low rank since each row of $Y$ can be represented by a linear combination of the rows in $[X; \mathrm{1}^T]$, i.e., $\mathrm{rank}([Y; X; 1^T]) = \mathrm{rank}([X; 1^T])$. Actually, the feature matrix $X$ can also be assumed to be of low rank, i.e., $\mathrm{rank}(X) \ll min(d,n)$. This assumption is reasonable since high dimensional data typically lies on a manifold, and the assumption succeed in many component analysis techniques, e.g., principal component analysis (PCA) and Fisher's linear discriminant analysis (FLDA).

Let $Z^0=[Y^0; X^0; \mathrm{1}^T]$ and $\Omega_X$, $\Omega_Y$ be the set of known entries in $X^0$ and $Y^0$ respectively. To recover the underlying feature matrix and predict the soft labels, Goldberg et al.~\cite{A-Goldberg-et-al-NIPS-2010} proposed to minimize the rank of $Z$, as well as the error between $Z$ and $Z^0$ for the entries in $\Omega_X$, $\Omega_Y$. Thus the optimization problem, MC-1, is
\begin{equation}
\label{eq:MC_1_Formulation}
\begin{split}
\mathop{\mathrm{argmin}}_Z &\ \mu \|Z\|_* + \frac{1}{|\Omega_X|} \sum_{i,j \in \Omega_X} c_x (z_{ij},z_{ij}^0) \\
& + \frac{\lambda}{|\Omega_Y|} \sum_{i,j \in \Omega_Y} c_y (z_{ij},z_{ij}^0), \\
\mathrm{s.t.} \ \ & \mathrm{z}_{(m+d+1)} = \mathbf{1}^T.
\end{split}
\end{equation}
where $\|Z\|_*$ is the nuclear norm (sum of singular values) of $Z$, $c_x$ and $c_y$ are the loss function for the features and labels respectively. Both $\mu$ and $\lambda$ are the trade-off parameters. In~\cite{A-Goldberg-et-al-NIPS-2010}, $c_x$ is chosen to be the least squares loss and $c_y$ is the log loss.

The problem (\ref{eq:MC_1_Formulation}) can be solved by a modified fixed point continuation (FPC) algorithm, which is to alternate between the gradient descent, $A^k = Z^k - \tau g(Z^k)$, and the shrinkage $Z^{k+1}=S_{\tau \mu} (A^k)$. The step size $\tau$ can be easily computed according to~\cite{A-Goldberg-et-al-NIPS-2010, R-Cabral-et-al-NIPS-2011}. Here, $k$ is the iteration step and $g(Z^k)$ is the matrix gradient given by
\begin{numcases}
{g(z_{ij})=}{}
\begin{split}
\label{eq:MC_1_Gradient}
\frac{\lambda}{|\Omega_Y|} \frac{-z_{ij}^0}{1+\mathrm{exp}(z_{ij}^0 z_{ij})}, & \ \mathrm{if} z_{ij} \in \Omega_Y \\
\frac{1}{|\Omega_X|} (z_{ij} - z_{ij}^0),\ \ \ \ \ & \ \mathrm{if} z_{ij} \in \Omega_X \\
0, \ \ \ \ \ \ \ \ \ \ \ \ \ \ \ & \ \mathrm{otherwise}.
\end{split}
\end{numcases}
However, images are usually represented by feature histograms, which consist of positive values naturally. The formulation (\ref{eq:MC_1_Formulation}) will introduce negative values to the features. Therefore, in~\cite{R-Cabral-et-al-NIPS-2011}, all feature vectors are constrained to be positive or belong to a simplex, and the corresponding MC-Pos and MC-Simplex algorithms are induced. Besides, it is suggested in~\cite{R-Cabral-et-al-NIPS-2011} to use the Pearson's $\chi^2$ distance for $c_x$ to take the asymmetry of the feature histogram into consideration. For the labels, the authors used a generalized version of the log loss, i.e., $c_y(u,v) = \frac{1}{\gamma} \mathrm{log}(1 + \mathrm{exp}(-\gamma u v))$, which is approximate to the hinge loss, and more suitable for classification than the original log loss.

\begin{figure*}
\centering
\includegraphics[width=1.6\columnwidth]{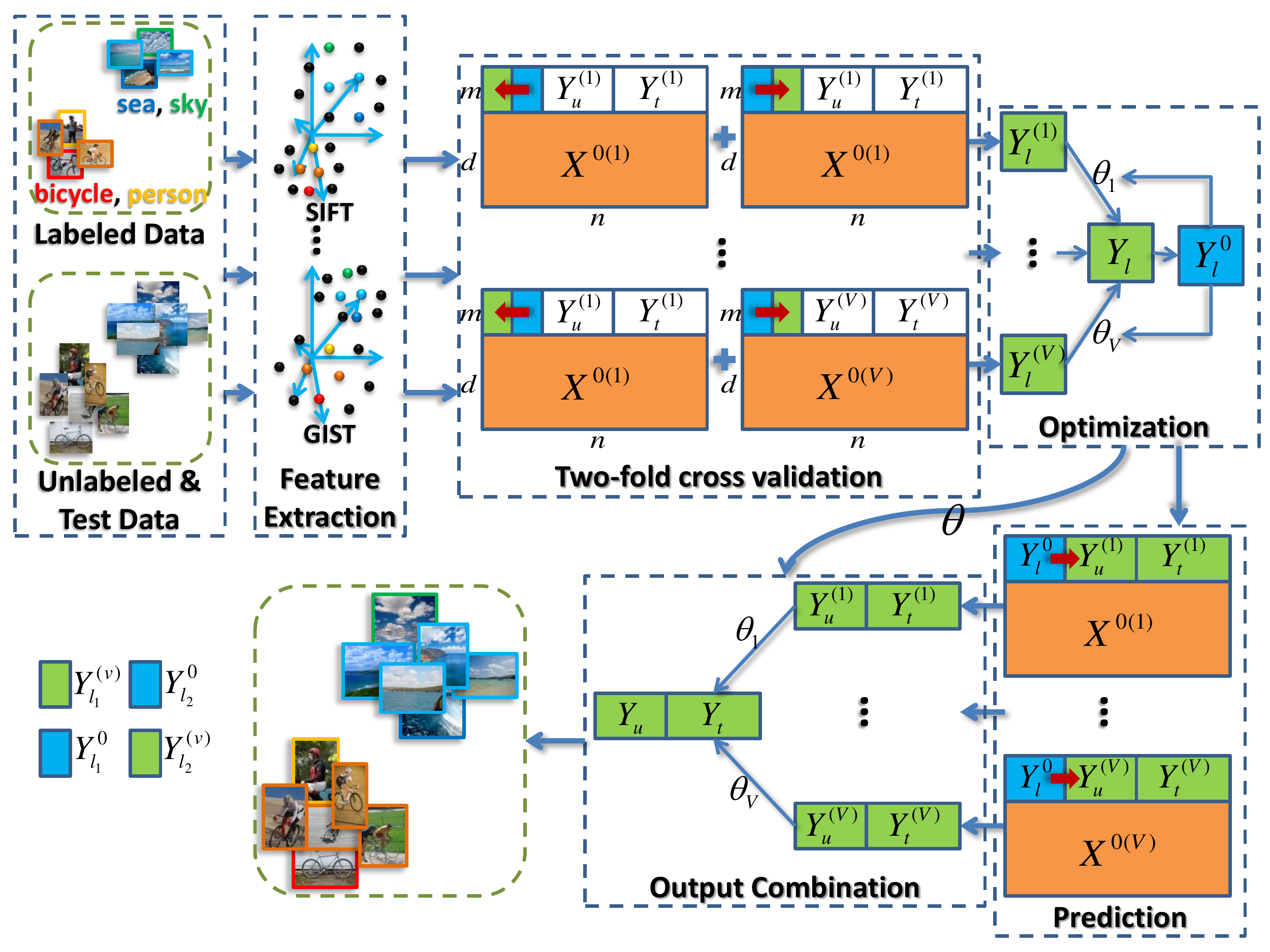} \\
\caption{System diagram of the multi-view matrix completion framework for transductive (semi-supervised) multi-label image classification. For all the data, features from different views (e.g. SIFT~\cite{D-Lowe-IJCV-2004}, GIST~\cite{A-Oliva-and-A-Torralba-IJCV-2001}, etc.) are extracted. Then a two-step algorithm is performed: 1) generate data for learning $\theta$ by performing a two-fold cross validation procedure on the labeled data set for each view; 2) learn $\theta$ with the constructed optimization problem and the generated data. Finally, for all the views, we predict the labels of the unlabeled and test data, and combine the outputs using the learned $\theta$.}
\label{fig:System_Diagram}
\end{figure*}

\section{Multi-view matrix completion}
\label{sec:Multiview_MC}

In this section, we first present our multi-view matrix completion (MVMC) framework, and then develop an algorithm that directly optimizes average precision for transductive (semi-supervised) multi-label image classification. The MVMC framework is depicted in Fig. \ref{fig:System_Diagram}. For all the labeled, unlabeled and test images, we extract different kinds of features, such as SIFT~\cite{D-Lowe-IJCV-2004} and GIST~\cite{A-Oliva-and-A-Torralba-IJCV-2001}. Then we construct the stacked matrix $Z^{(v)}$ for the $v$'th view. To combine the multiple views for matrix completion, a natural idea is to weightedly sum the different feature matrices $X^{(v)}, v=1, \ldots ,V$, where $V$ is the number of views. However, the dimensionality of different views varies. Although we can utilize some dimensionality reduction algorithm, such as kernel PCA (KPCA), to preprocess the features, the summation of different matrices lacks physical interpretation. Therefore, we propose to combine the output label matrices of different views. KPCA is still employed to preprocess the features to significantly reduce the time complexity of the MC algorithm. The MC-1 algorithm is adopted to complete the matrix for each view since the processed features can be either positive or negative. We use the least squares loss for $c_x$ and the generalized log loss (a smooth approximation of the hinge loss) for $c_y$. Subsequently, we proposed a two-step algorithm to learn the output combination coefficients of the different views: \\
$\bullet$ \textbf{Generate training data for each view.} The training data we referred here is not the features to complete the matrix $Z^{(v)}$, but the output labels to learn the combination $\theta$. We generate the data by assuming parts (e.g. half) of the labeled data as unlabeled, and predict their labels using the other parts. Such a process is then conducted conversely. In this way, we obtain the predicted labels $Y_l^{(v)}, v=1, \ldots ,V$ of the labeled data, whose labels $Y_l^0$ are actually known. \\
$\bullet$ \textbf{Learn the combination coefficients.} In our formulation, the final output is a linear combination of the multiple outputs obtained from different views. Thus, we use the weighted summation $Y_l = \sum_{v=1}^V \theta_v Y_l^{(v)}$ to approximate the ground-truth $Y_l^0$. By minimizing the approximation error, we can learn the weight $\theta$.

Finally, for each view, we predict the labels of the unlabeled and test data by utilizing all the labeled data. The multiple predictions are combined with the learned $\theta$.

\subsection{The general formulation for learning $\theta$}

To simplify the derivation, we convert the original two dimensional index $(i,j)$ of a sample to one dimensional $k$ by the use of vectorization. Let us define $\Omega_{Y_l}$ as the set of all the known labels' entries, and $\{p_k^{(v)}\}, k \in \Omega_{Y_l}$ as the set of generated data (predicted values of the labeled data) from the $v$'th view. Thus our prediction function can be written as $f(\mathrm{p}_k) = \sum_{v=1}^V \theta_v p_k^{(v)} = \theta^T \mathrm{p}_k$, where $\mathrm{p}_k = [p_k^{(1)}, \ldots ,p_k^{(V)}]^T$ is the predictions of all the views for the $k$'th sample. Then we have the following optimization problem,
\begin{equation}
\label{eq:MVMC_General_Formulation}
\begin{split}
\mathop{\mathrm{argmin}}_\theta &\ \frac{1}{2N} \sum_{k=1}^N L\left(f(\mathrm{p}_k),y_k^0\right) + \frac{\eta}{2} \|\theta\|_2^2, \\
\mathrm{s.t.} &\ \sum_{v=1}^V \theta_v=1, \theta_v \geq 0,
\end{split}
\end{equation}
where $N = |\Omega_{Y_l}|$ is the number of entries in $\Omega_{Y_l}$, which is equal to $n_l \times m$. Here, $n_l$ is the number of labeled samples. The regularization term $\|\theta\|_2^2$ is used to control the model complexity, and $\eta$ is the trade-off parameter. Here, $L$ is some pre-defined convex loss. It is easy to verify that the problem (\ref{eq:MVMC_General_Formulation}) is convex, and thus we can obtain the global solution. In this paper, we first choose $L$ to be the least squares (LS) loss, which will lead to a quite efficient solution.

\subsection{A least squares formulation of MVMC (MVMC-LS)}

If we choose $L$ to be the least squares loss, i.e. $L(f(x),y)=(f(x)-y)^2$, then the optimization problem becomes
\begin{equation}
\label{eq:MVMC_LS_Formulation}
\begin{split}
\mathop{\mathrm{argmin}}_\theta &\ \frac{1}{2N} \sum_{k=1}^N \left((\theta^T p_k)-y_k^0\right)^2 + \frac{\eta}{2} \|\theta\|_2^2, \\
\mathrm{s.t.} &\ \sum_{v=1}^V \theta_v=1, \theta_v \geq 0,
\end{split}
\end{equation}
Let $P=[p_1,\ldots,p_N] \in \mathbb{R}^{V \times N}$ and $\mathrm{y}^0=[y_1^0,\ldots,y_N^0]^T$. The objective of the above problem can be written in a compact form as $\frac{1}{2N} \|P^T \theta - \mathrm{y}^0\|_F^2 + \frac{\eta}{2} \|\theta\|_2^2$, and then the problem (\ref{eq:MVMC_LS_Formulation}) can be reformulated as
\begin{equation}
\label{eq:MVMC_LS_Conpact_Form}
\begin{split}
\mathop{\mathrm{argmin}}_\theta &\ \frac{1}{2} (\theta^T H \theta - 2\theta^T \mathrm{h}) + \frac{\eta}{2} \|\theta\|_2^2, \\
\mathrm{s.t.} &\ \sum_{v=1}^V \theta_v=1, \theta_v \geq 0,
\end{split}
\end{equation}
where the constant term $\frac{1}{N} (\mathrm{y}^0)^T \mathrm{y}^0$ has been omitted, $H=\frac{1}{N} P P^T$ is a $V \times V$ semi-definite and usually full-rank matrix, and $h=[h_1,\ldots,h_V]^T$ with each $h_v=\frac{1}{N} P_{(v)}^T \mathrm{y}^0$. Here, $P_{(v)}$ denotes the $v$'th row of the matrix $P$, that is the prediction from the $v$'th view. To solve this problem, the coordinate descent algorithm is adopted. In each round of iteration, only two variables $\theta_i$ and $\theta_j$ are selected to update, while the others are fixed. We firstly only consider the constraint $\sum_{v=1}^V \theta_v=1$. By using the Lagrange method and note that $\theta_i+\theta_j$ do not change in the iteration, we have the following solution for updating $\theta_i$ and $\theta_j$:
\begin{numcases}{}
\label{eq:Update_Rule_Theta}
\begin{split}
& \theta_i^* = \frac{\eta(\theta_i+\theta_j)+(h_i-h_j)+\varepsilon_{ij}}{(H_{ii}-H_{ij}-H_{ji}+H_{jj})+2\eta}, \\
& \theta_j^* = \theta_i+\theta_j-\theta_i^*,
\end{split}
\end{numcases}
where $\varepsilon_{ij}=(H_{ii}-H_{ij}-H_{ji}+H_{jj})\theta_i-\sum_k(H_{ik}-H_{jk})\theta_k$. By further taking the constraint $\theta_v \geq 0$ into consideration, we have
\begin{numcases}{}
\notag
\begin{split}
& \theta_i^*=0,\theta_j^*=\theta_i+\theta_j, \mathrm{if}\ \eta(\theta_i+\theta_j)+(h_i-h_j)+\varepsilon_{ij} \leq 0, \\
& \theta_j^*=0,\theta_i^*=\theta_i+\theta_j, \mathrm{if}\ \eta(\theta_i+\theta_j)+(h_j-h_i)+\varepsilon_{ji} \leq 0.
\end{split}
\end{numcases}
From solution (\ref{eq:Update_Rule_Theta}), we can see that larger $h_i$ implies larger $\theta_i$. This is because $h_i=\frac{1}{N} P_{(i)}^T \mathrm{y}^0$ denotes the similarity between the prediction of the $i$'th view and the ground-truth.

In spite of the efficiency of the LS formulation, the LS loss is designed to optimize the accuracy performance, which is not appropriate for multi-label classification \cite{Pascal-VOC, M-Guillaumin-et-al-CVPR-2010}. Thus the obtained solution may be unsatisfactory. The hinge loss used in support vector machine (SVM) is not adopted for the same reason. Besides, the least squares and hinge loss are not robust loss functions \cite{A-Christmann-and-I-Steinwart-JSTOR-2007, YC-Wu-and-YF-Liu-JASA-2007}. Therefore, we propose to directly optimize the average precision (AP), which is a critical criterion for evaluating the multi-label classification performance, and we can prove that the algorithm that utilizes the AP loss is more robust than that adopts the least squares or hinge loss. To this end, better view combination coefficients $\{\theta_v\}$ can be found hopefully. In the following, we first present the AP formulation of MVMC, and then give some theoretical analysis of the proposed algorithm, i.e., the robustness of the algorithm that adopts the AP loss compared with the least squares and hinge loss.


\subsection{Optimizing average precision in MVMC (MVMC-AP)}

Different from the least squares formulation, where the loss is point-wise and calculated on two scalar elements, the average precision (AP) loss is computed over two vectors for each label (category) and thus list-wise. Before presenting the AP loss, we first introduce the AP score. Suppose the input space is $\mathcal{C}$ and the output space is $\mathcal{O}$ (rankings\footnote{It should be noted that the ranking we refer to here is an ordered sequence and the rank value is a number in the sequence, not the notation ``rank'' of a matrix we used in Section \ref{sec:Transduction_with_MC}.} over a corpus $\mathcal{S} = \{d_1, \ldots ,d_{|\mathcal{S}|} \}$, each $d_i$ is a sample). In this paper (transductive multi-label classification), $\mathcal{C}$ consists of the different labels (categories), which correspond to the possible queries in information retrieval \cite{C-Manning-et-al-Cambridge-Book-2008}. For a certain label, if $\hat{o}$ is a prediction vector for the samples in $\mathcal{S}$ and $o$ is the corresponding ground-truth, then the AP score can be defined as
\begin{equation}
\mathrm{AP}(r(o),r(\hat{o})) = \frac{1}{N_{pos}} \sum_k Prec@k,
\end{equation}
where $r(\cdot)$ is a vector of rank values. The ground-truth ranking $r(o)$ has only two rank values, i.e., $1$ for the positive samples and $0$ otherwise. The prediction ranking $r(\hat{o})$ is the sorting result of the predictions in $\hat{o}$, a larger prediction value corresponding to a higher rank value. Here, $N_{pos} = |\{i:r_i(o)=1\}|$ is the total number of positive samples, and $Prec@k$ is the percentage of positive samples in the top $k$ samples, where the samples in the corpus are assumed to have been sorted according to the prediction $\hat{o}$. 

\begin{table}[!t]
\setlength\tabcolsep{2pt}
\renewcommand{\arraystretch}{1.3}
\caption{A toy example of the predictions and rankings}
\label{tab:Toy_Example_Rank}
\centering
\begin{tabular}{c|cccccccc}
\hline
$r(o)$ & 1 & 0 & 0 & 0 & 1 & 0 & 0 & 0 \\
\hline \hline
$\hat{o}_1$ & 0.4 & 0.3 & 0.2 & 0.1 & -0.1 & -0.2 & -0.3 & -0.4 \\
\hline
$r(\hat{o}_1)$ & 8 & 7 & 6 & 5 & 4 & 3 & 2 & 1 \\
\hline
$\hat{o}_2$ & -0.4 & -0.3 & -0.2 & -0.1 & 0.1 & 0.2 & 0.3 & 0.4 \\
\hline
$r(\hat{o}_2)$ & 1 & 2 & 3 & 4 & 5 & 6 & 7 & 8 \\
\hline
\end{tabular}
\end{table}

Table \ref{tab:Toy_Example_Rank} is a toy example of two predictions. The AP scores are calculated as $\mathrm{AP}(r(o),r(\hat{o}_1))=\frac{1}{2} (1/1+2/5)=0.7$ and $\mathrm{AP}(r(o),r(\hat{o}_2))=\frac{1}{2} (1/4+2/8)=0.25$ respectively. However, the accuracy of the two predictions are both $0.5$ if we choose the threshold to be zero, and classify the samples predicted to be higher than the threshold as positive. We find that the accuracy is not appropriate in this case since the first classifier assigns much higher rank values to the positive samples than the second classifier, and thus performs much better. Such cases are common in multi-label classification since there are usually much more negative samples than the positive ones for each label. This is an intuition of why adopting the AP loss here, which is defined as
\begin{equation}
\label{eq:Average_Precision_Loss}
\Delta_{ap} (o,\hat{o}) = 1 - \mathrm{AP}(r(o),r(\hat{o})),
\end{equation}
Usually, the mean of the AP scores of all labels, i.e., mAP is adopted for evaluation. Therefore, the loss calculation will be preformed over all labels. In the following, we show how to incorporate the AP loss into an optimization problem for learning $\theta$.

The central idea of optimizing AP in MVMC is to transform the multi-label classification into a retrieval problem, and regard each label as a query. Given a certain label $t$, the aim is to find a ranking $o$ that maximizes the discriminant function:
\begin{equation}
\label{eq:Discriminant_Function}
F(t,o,\theta) = \theta^T \Psi(t,o),
\end{equation}
which is assumed to be linear (parameterized by $\theta$) in some combined feature representation given by:
\begin{equation}
\label{eq:Combined_Feature_Representation}
\Psi(t,o) = \frac{1}{|\mathcal{S}^t| \cdot |\bar{\mathcal{S}}^t|} \sum_{i:d_i \in \mathcal{S}^t} \sum_{j:d_j \in \bar{\mathcal{S}}^t} \left[ o_{ij} (\phi(t,d_i) - \phi(t,d_j)) \right],
\end{equation}
where $d_i$ and $d_j$ are the samples, $\mathcal{S}^t$ and $\bar{\mathcal{S}}^t$ denote the set of positive and negative samples of $S$ for label $t$. Here, the pairwise orderings is utilized, i.e., $\mathcal{O} \subset \{-1,0,+1\}^{|\mathcal{S}| \times |\mathcal{S}|}$. For each $o \in \mathcal{O}$, $o_{ij} = +1$ if $d_i$ is ranked ahead of $d_j$, $o_{ij} = 0$ if $d_i$ and $d_j$ have equal rank, and $o_{ij} = -1$ otherwise. We assume that the rankings is complete, i.e., $o_{ij}$ is either $+1$ or $-1$ (never $0$).

We can predict a ranking (of the samples) for label $t$ with a learned $\theta$. However, this is not the point of this paper and we only concentrate on learning the weight vector $\theta$. Here, we define the feature mapping function $\phi$ as $\phi(t,d) = p$. Each element of $p$ corresponds to the prediction of a certain view, and thus $\theta^T \phi(t,d)$ is a combined prediction of different views. For different labels, the feature mappings are the same, only the ground-truth orderings change. In this way, we can learn $\theta$ to combine different views by the use of the average precision (AP) loss.

Following \cite{Y-Yue-et-al-SIGIR-2007}, we use the structural SVM formulation to learn $\theta$, where an additional simplex constraint is added:
\begin{equation}
\label{eq:Structual_SVM_Formulation}
\begin{split}
\arg\min
& \frac{1}{2} \| \theta \|^2 + C \sum_{t=1}^m \xi_t, \\
\mathrm{s.t.} \ \ \
& \forall t, \forall o \in \mathcal{O} \backslash o_t: \theta^t \delta \Psi_t(o) \geq \Delta(o_t,o) - \xi_t, \xi_t \geq 0, \\
& \sum_v \theta_v = 1, \theta_v \geq 0,
\end{split}
\end{equation}
where $o_t$ is the ground-truth ranking for the label $t$. $C = \frac{1}{2N\eta}$, and we have defined $\delta \Psi_t(o) = \Psi(t,o_t) - \Psi(t,o)$. We propose to solve the problem (\ref{eq:Structual_SVM_Formulation}) using an alternating algorithm in the dual formulation. Note that for $\theta_v \geq 0,v=1, \ldots ,V$, the constraint $\sum_{v=1}^V \theta_v = 1$ can be satisfied by a simple normalization, so we left this sum-to-one constraint to be considered later. By introducing the Lagrangian, we obtain
\begin{equation}
\label{eq:Lagrangian_Form}
\begin{split}
\mathcal{L}(\theta, \xi, \alpha, \beta, \zeta) = & \frac{1}{2} \| \theta \|^2 + C \sum_{t=1}^m \xi_t - \sum_t \beta_t \xi_t - \sum_v \zeta_v \theta_v \\
& - \sum_{t,o \neq o_t} \alpha_{to} \left( \theta^T \delta \Psi_t(o) - \Delta(o_t,o) + \xi_t \right),
\end{split}
\end{equation}
Taking the partial derivatives of $L$ w.r.t. $\theta$, $\xi_t$ and setting them to be zero,
\begin{equation}
\notag
\begin{split}
\frac{\partial \mathcal{L}}{\partial \theta} & = 0, \Rightarrow \theta = \sum_{t, o \neq o_t} \alpha_{to} \delta \Psi_t(o) + \zeta, \\
\frac{\partial \mathcal{L}}{\partial \xi_t} & = 0, \Rightarrow C - \alpha_{to} - \beta_t = 0.
\end{split}
\end{equation}
where $\zeta = [\zeta_1, \ldots ,\zeta_V]^T$ and the second identity is equivalent to $0 \leq \alpha_{to} \leq C$ since the Lagrange multiplier $\beta_t \geq 0$. By substituting $\theta$ back into (\ref{eq:Structual_SVM_Formulation}), we obtain the dual
\begin{equation}
\label{eq:Dual_Formulation}
\begin{split}
\arg\max_{\alpha,\zeta}
& \sum_{t, o \neq o_t} \alpha_{to} \Delta(o_t,o) - \frac{1}{2} \sum_{t,o \neq o_t} \sum_{s,\bar{o} \neq o_s} \alpha_{to} \alpha_{s\bar{o}} K_{(to)(s\bar{o})} \\
& - \zeta^T \sum_{t,o \neq o_t} \alpha_{to} \delta \Psi_t(o) - \frac{1}{2} \zeta^2, \\
\mathrm{s.t.} \ \ \
& 0 \leq \alpha_{to} \leq C, \zeta_v \geq 0.
\end{split}
\end{equation}
where $K_{(to)(s\bar{o})} = \langle \delta \Psi_t(o), \delta \Psi_s(\bar{o}) \rangle$. Let $\alpha = [\alpha_{1,o \neq o_1}^T, \ldots ,\alpha_{m,o \neq o_m}^T]^T \in \mathbb{R}^{m|\mathcal{S}|-m}, \Delta = [\Delta^T (o_1,o \neq o_1), \ldots ,\Delta^T(o_m, o \neq o_m)]^T$ and $\delta \Psi = [\delta \Psi_1(o \neq o_1), \ldots ,\delta \Psi_m(o \neq o_m)]$ with each $\delta \Psi_t(o \neq o_t) \in \mathbb{R}^{V \times (|\mathcal{S}|-1)}$, the problem (\ref{eq:Dual_Formulation}) can be written in a compact form as
\begin{equation}
\label{eq:Compact_Form}
\begin{split}
\arg\max_{\alpha, \zeta}
& \Delta^T \alpha - \frac{1}{2} \alpha^T K \alpha - \zeta^T \delta \Psi \alpha - \frac{1}{2} \zeta^T \zeta, \\
\mathrm{s.t.} \ \ \
& 0 \leq \alpha_{to} \leq C, \zeta_v \geq 0.
\end{split}
\end{equation}
We propose to solve this problem using the alternating optimization strategy. For fixed $\zeta$, the problem (\ref{eq:Compact_Form}) becomes
\begin{equation}
\label{eq:Alpha_Optimization}
\begin{split}
\arg\max_{\alpha}
& (\Delta^T - \zeta^T \delta \Psi) \alpha - \frac{1}{2} \alpha^T K \alpha, \\
\mathrm{s.t.} \ \ \
& 0 \leq \alpha_{to} \leq C,
\end{split}
\end{equation}
which is a structural SVM \cite{I-Tsochantaridis-et-al-JMLR-2005} formulation with the linear part $(\Delta^T - \zeta^T \delta \Psi) \alpha$. A cutting-plane algorithm is introduced to solve this problem and the most violated constraint can be found using the algorithm presented in \cite{Y-Yue-et-al-SIGIR-2007}. For fixed $\alpha$, the problem (\ref{eq:Compact_Form}) can be reformulated as
\begin{equation}
\label{eq:Zeta_Optimization}
\begin{split}
\max_{\zeta}
& (-\alpha^T \delta \Psi^T) \zeta - \frac{1}{2} \zeta^T \zeta, \\
\mathrm{s.t.} \
& \zeta_v \geq 0.
\end{split}
\end{equation}
This is a quadratic programming (QP) problem and can be solved quite efficiently using a standard SVM solver. It can be easily verified that the Hessian matrix of (\ref{eq:Compact_Form}) $H_e(\alpha, \zeta) = -\left[ \begin{array}{cc} K & \delta \Psi^T \\ \delta \Psi & I_V \end{array} \right]$ is negative semi-definite, where $I_V$ is the $V \times V$ identity matrix, and thus the problem (\ref{eq:Compact_Form}) is jointly concave w.r.t. $\alpha$ and $\zeta$. Besides, the sub-problems (\ref{eq:Alpha_Optimization}) and (\ref{eq:Zeta_Optimization}) are concave w.r.t. $\alpha$ and $\zeta$ respectively. Therefore, by alternatively solving (\ref{eq:Alpha_Optimization}) and (\ref{eq:Zeta_Optimization}), the algorithm will converge to the global solution of (\ref{eq:Compact_Form}).


\subsection{Complexity analysis}

The complexity of MVMC has two parts: the first is determined by the MC-based classification of each view, and the second is determined by the learning of the view combination coefficients. In this paper, the MC-based classification is carried out by optimizing the MC-1 problem (\ref{eq:MC_1_Formulation}). We can reduce the complexity of the MC-based classification algorithm MC-1 \cite{R-Cabral-et-al-NIPS-2011} to $¦¯(st(m+\bar{d}+n))$ by exploiting the approximate SVD based fixed point continuation (FPCA) \cite{SQ-Ma-et-al-MP-2011} algorithm, where $m$ and $n$ are the number of class labels and all samples respectively, $\bar{d}$ is the average feature dimension, $t$ is the iteration numbers, and $s$ is the number of elements in the $\mu$ sequence of the continuation step. In common, $s < 50$ and $t < 100$. Thus, the adopted MC-1 can solve large matrix rank minimization problems efficiently.

With regard to the learning of the combination coefficients ${\theta_v}_{v=1}^V$, we use the weighted summation $Y_l = \sum_{v=1}^V \theta_v Y_l^{(v)}$ to approximate the groundtruth $Y_l^0$. Weights $\{\theta_v\}_{v=1}^V$ are obtained by minimizing the approximation error. The size of each $Y_l^{(v)}$ is $n_l \times m$, where $n_l$ is the number of labeled data. This means that the complexity of the proposed view combination procedure is not dependent on the amount of all samples, but only on the labeled sample size, which is usually small in transductive (semi-supervised) classification. For MVMC-LS, the time complexity is $¦¯(V^2 (n_l \times m))$, where $V$ is the number of views and is usually smaller than $10$. For MVMC-AP, structure SVM is exploited for optimization. According to \cite{T-Joachims-et-al-MLJ-2009}, the time complexity of the cutting-plane method used in structural SVMs is linear in the number of training samples. Thus the computational time cost of learning the view combination coefficients is $¦¯(T(n_l \times m))$, where $T$ is the number of iterations and is independent on the number of samples \cite{T-Joachims-et-al-MLJ-2009}. To this end, the view combination procedure is also very efficient, and according to our experience, it is more efficient than the MC-based classification.

Therefore, the time complexity of MVMC-LS and MVMC-AP are $¦¯(Vst(m+\bar{d}+n) + V^2(n_l \times m))$ and $¦¯(Vst(m+\bar{d}+n) + T(n_l \times m))$, respectively. The former is often more efficient since $V$ is usually small.

\subsection{Robustness analysis}

In this section, we aim to prove that AP loss is more robust than the least squares and hinge loss when learning $\theta$ for MVMC. That is, AP is more tolerate of noise. We present the definition of the robustness here for completeness.
\begin{defn}
\label{def:Robustness}
(Robustness~\cite{H-Xu-and-S-Mannor-MLJ-2012}) Let $\mathcal{Z}$ be the set that each sample is drawn from, and $\mathbf{s}$ be a training sample set $\{s_1, \ldots, s_n\}$. A learning algorithm $\mathcal{A}$ is $(K, \epsilon(\cdot))$ robust, for $K \in \mathbb{N}$ and $\epsilon(\cdot): \mathcal{Z}^n \mapsto \mathbb{R}$, if $\mathcal{Z}$ can be partitioned into $K$ disjoint sets, denoted by $\{\mathcal{C}_i \}_{i=1}^K$, such that the following holds for all $\mathbf{s} \in \mathcal{Z}^n$:
\begin{equation}
\begin{split}
& \forall s \in \mathbf{s}, \forall z \in \mathcal{Z}, \forall i = 1, \ldots, K: \\
& \mathrm{if} s,z \in \mathcal{C}_i, \mathrm{then} |l(h_\mathbf{s},s) - l(h_\mathbf{s},z)| \leq \epsilon(\mathbf{s}),
\end{split}
\end{equation}
where $h_\mathbf{s}$ represents the hypothesis learned using $\mathcal{A}$ on the training set $\mathbf{s}$. The sets $\{\mathcal{C}_i\}_{i=1}^K$ can be regarded as sets with elements having some similarity to each other, such as the distance.
\end{defn}

Let $\mathcal{A}_\mathrm{ap}$, $\mathcal{A}_\mathrm{ls}$ and $\mathcal{A}_\mathrm{svm}$ denote the learning algorithms which adopt the AP loss, least squares loss and hinge loss in (\ref{eq:MVMC_General_Formulation}) respectively.  It is well known that the ranking based criteria are critical for multi-label classification \cite{K-Aas-and-L-Eikvil-TR-Norwegian-1999, ML-Zhang-and-ZH-Zhou-PR-2007, M-Guillaumin-et-al-CVPR-2010, HQ-Minh-and-V-Sindhwani-ICML-2011}, and we can also prove that $\mathcal{A}_\mathrm{ap}$ is more robust than $\mathcal{A}_\mathrm{ls}$ and $\mathcal{A}_\mathrm{svm}$ when learning $\theta$ for MVMC. This is easy to understand since the AP loss is list-wise, while the LS and hinge loss are both point-wise.
\begin{thm}
\label{thm:Robustness}
We can partition the sample space $\mathcal{Z}$ into $N$ disjoint sets $\{\mathcal{C}_i\}_{i=1}^N$ such that $\mathcal{A}_\mathrm{ap}$ is $(N,0)$ robust, $\mathcal{A}_\mathrm{ls}$ is $(N,a)$ robust and $\mathcal{A}_\mathrm{svm}$ is $(N,b)$ robust, where $N$ is a positive integer and $a, b \neq 0$. That is, the $\mathcal{A}_\mathrm{ap}$ will tolerate to the noise which do not change the partition of the disjoint sets $\{\mathcal{C}_i\}_{i=1}^N$, while the $\mathcal{A}_\mathrm{ls}$ and $\mathcal{A}_\mathrm{svm}$ will not.
\end{thm}

\begin{remk}
\label{remk:Robustness}
It can also be seen from a general perspective that $\mathcal{A}_\mathrm{ap}$ is more robust than $\mathcal{A}_\mathrm{ls}$ and $\mathcal{A}_\mathrm{svm}$. That is, for any partitions, the least squares and hinge loss based algorithms cannot be $(K,0)$ robust, where $K \in \mathbb{N}$, while we can find at least one set of partition in which the AP loss based algorithm is $(K,0)$ robust.
\end{remk}

\begin{remk}
\label{remk:N_Zero_Robust}
In Theorem \ref{thm:Robustness}, the claim that $\mathcal{A}_\mathrm{ap}$ is $(N,0)$ robust does not mean that for all instances in the training set, the corresponding objective function losses are equal to each other. Instead, $\mathcal{A}_\mathrm{ap}$ is $(N,0)$ robust with respect to a particular partition $\{\mathcal{C}_i\}_{i=1}^N$, and such an assertion means that for all instances in a fixed set $\mathcal{C}_i, i \in \{1,\ldots,N\}$, the objective function losses are equal to each other. The central idea of Theorem \ref{thm:Robustness} is that, as we stated in the proof, for every point $P$ in a fixed prediction space $\mathcal{P}$, there has been a finite number of different rankings. Thus, we can partition the prediction space into a finite number of sets $\{\mathcal{C}_i\}_{i=1}^N$ such that all points in one set $\mathcal{C}_i$ share the same ranking. This is why the objective function losses for all instances in a fixed set are equal to each other and $\mathcal{A}_\mathrm{ap}$ is $(N,0)$ robust. However, $\mathcal{A}_\mathrm{ls}$ and $\mathcal{A}_\mathrm{svm}$ cannot be $(N',0)$ robust, where $N'$ is a finite counting number. This is because the corresponding losses have infinitely different values and we could never find a finite number of sets such that all points in one particular set share the same loss value.
\end{remk}

The following two lemmas are useful for proving Theorem \ref{thm:Robustness}.
\begin{lema}
\label{lema:N_Robust}
(\cite{H-Xu-and-S-Mannor-MLJ-2012}) Fix $\gamma > 0$ and metric $\rho$ of $\mathcal{Z}$, If algorithm $\mathcal{A}$ satisfies
\begin{equation}
\begin{split}
& |l(\mathcal{A}_{\mathcal{Z}^n}, z_1) - l(\mathcal{A}_(\mathcal{Z}^n), z_2)| \leq \epsilon(\mathcal{Z}^n), \\
& \forall z_1,z_2 \in \mathcal{Z}^n, \rho(z_1,z_2) \leq \gamma,
\end{split}
\end{equation}
and $N(\gamma/2, \mathcal{Z}, \rho) \leq \infty$, then $\mathcal{A}$ is $(N(\gamma/2, \mathcal{Z}, \rho), \epsilon(\mathcal{Z}^n))$ robust.
\end{lema}
\begin{lema}
\label{lema:Banach_Subset}
(\cite{T-Poggio-and-CR-Shelton-AMS-2002, A-Maurer-and-M-Pontil-TIT-2010}) Let $B$ be a ball of radius $r$ in an $N$-dimensional Banach space and $\epsilon > 0$. There exists a subset $B_\epsilon \subset B$ such that $|B_\epsilon| \leq (4r/\epsilon)^N$ and $\forall z \in B$, $\exists z' \in B_\epsilon$ with $\rho(z,z') \leq \epsilon$, where $\rho$ is the metric of the Banach space.
\end{lema}

We defer the detailed proof of Theorem \ref{thm:Robustness} in the last section. 

\section{Experiments}
\label{sec:Experiments}

In the experimental evaluation, we firstly compare the proposed MVMC-LS and MVMC-AP algorithms with the following MC-based strategies: 1) BMC, which is to use a single view that performs the best in MC; 2) CMC, which is to simply concatenate features of the multiple views in MC; 3) AMC, which is to average the MC outputs of the multiple views. Then we show the view combination coefficients learned by our MVMC-AP method. Finally, we compare MVMC-AP with some popular and competitive feature-level fusion \cite{A-Rakotomamonjy-et-al-JMLR-2008, M-Kloft-et-al-JMLR-2011}, and classifier-level fusion \cite{J-Kludas-et-al-AMR-2008} algorithms, as well as some competitive multi-label \cite{L-Sun-et-al-TPAMI-2011} and recently proposed semi-supervised multi-label classification approaches \cite{B-Wang-et-al-ICCV-2013}. Before all of these evaluations, we present the datasets and features we used, as well as our experimental settings.

\begin{figure*}
\centering
\subfigure{\includegraphics[width=0.65\columnwidth]{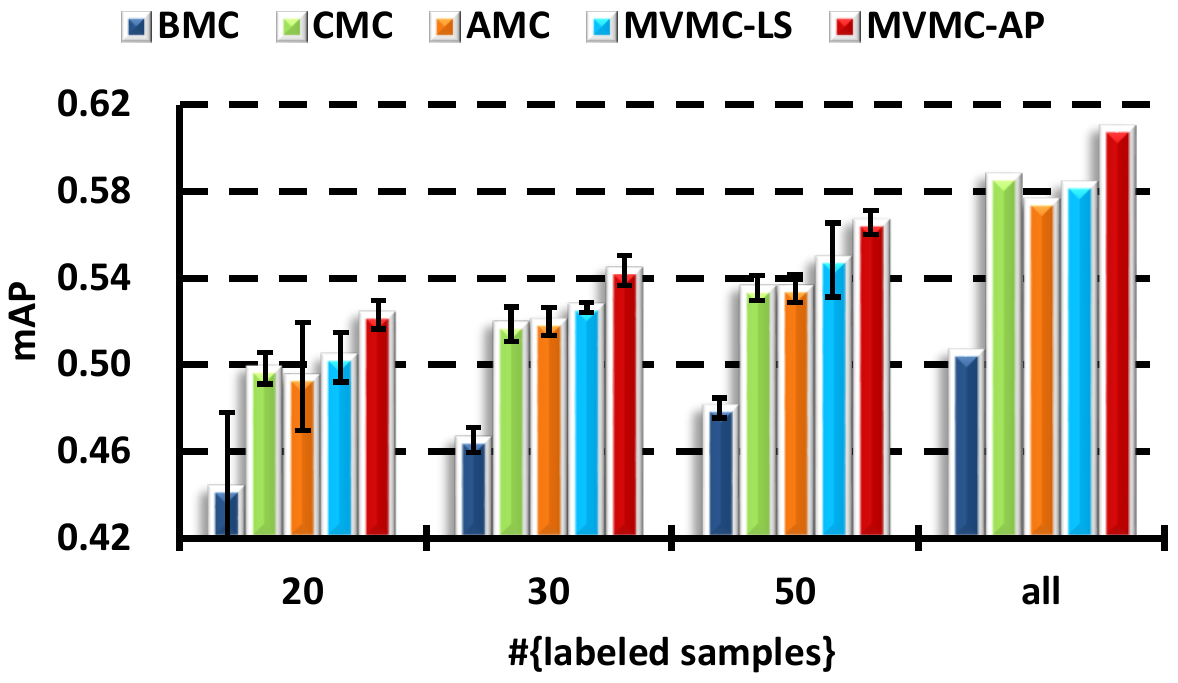}
}
\hfil
\subfigure{\includegraphics[width=0.65\columnwidth]{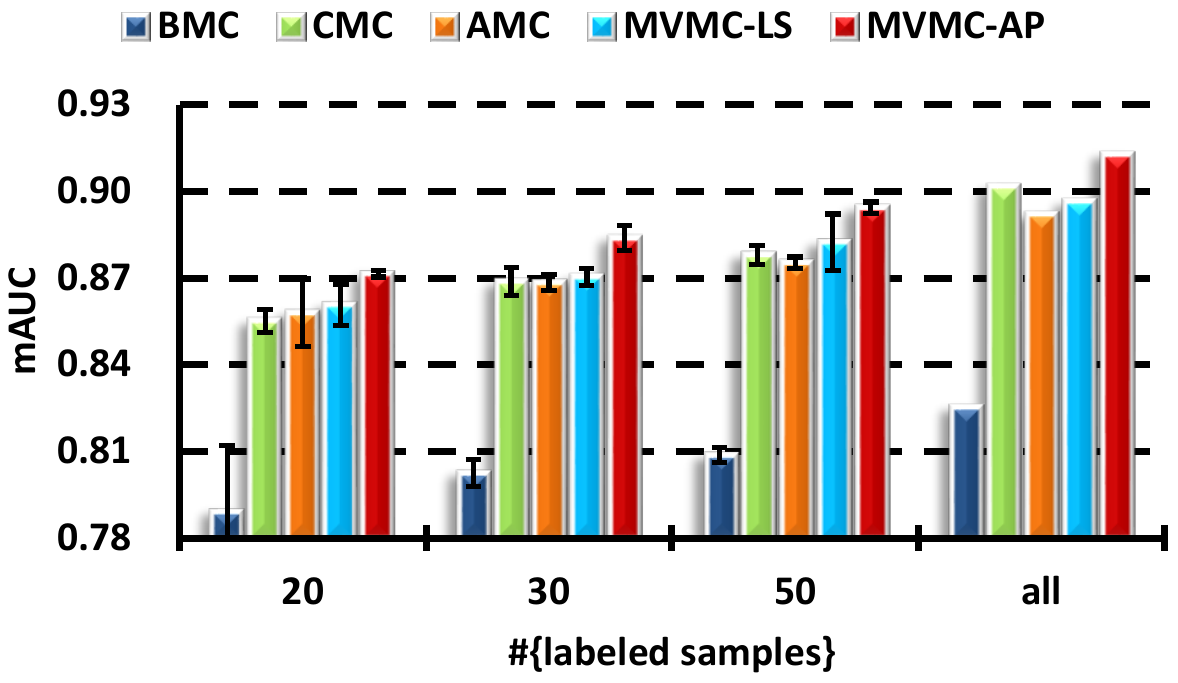}
}
\hfil
\subfigure{\includegraphics[width=0.65\columnwidth]{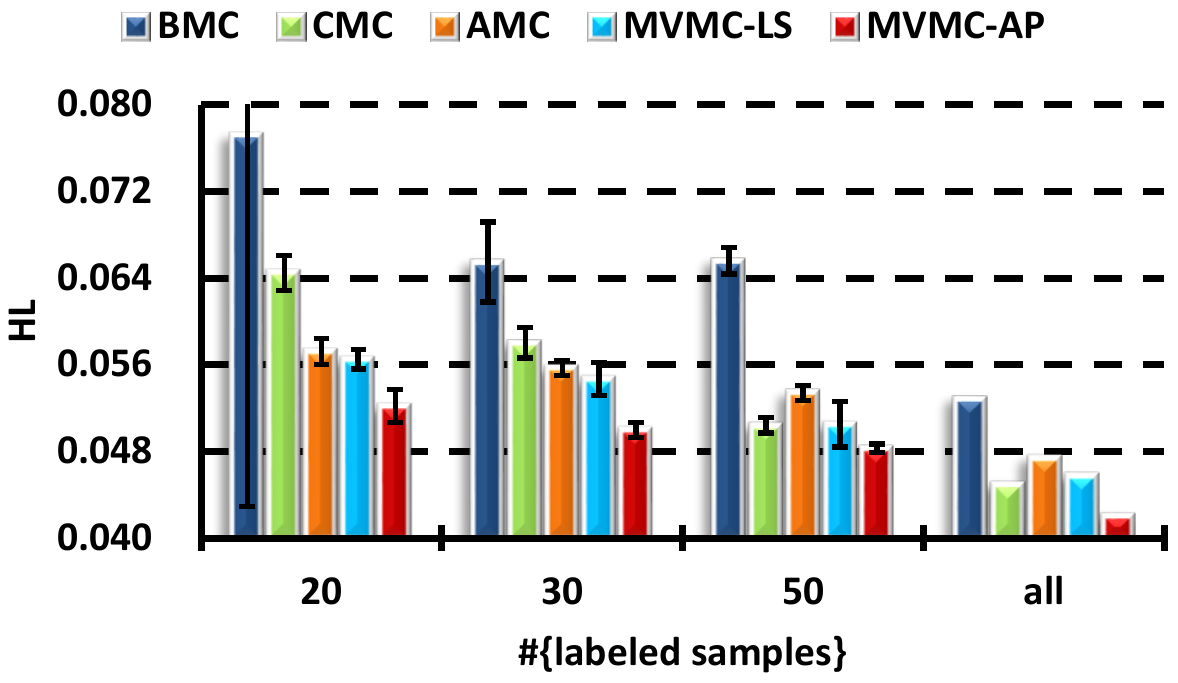}
}
\setcounter{subfigure}{0}
\hfil
\subfigure[mAP $\uparrow$]{\includegraphics[width=0.65\columnwidth]{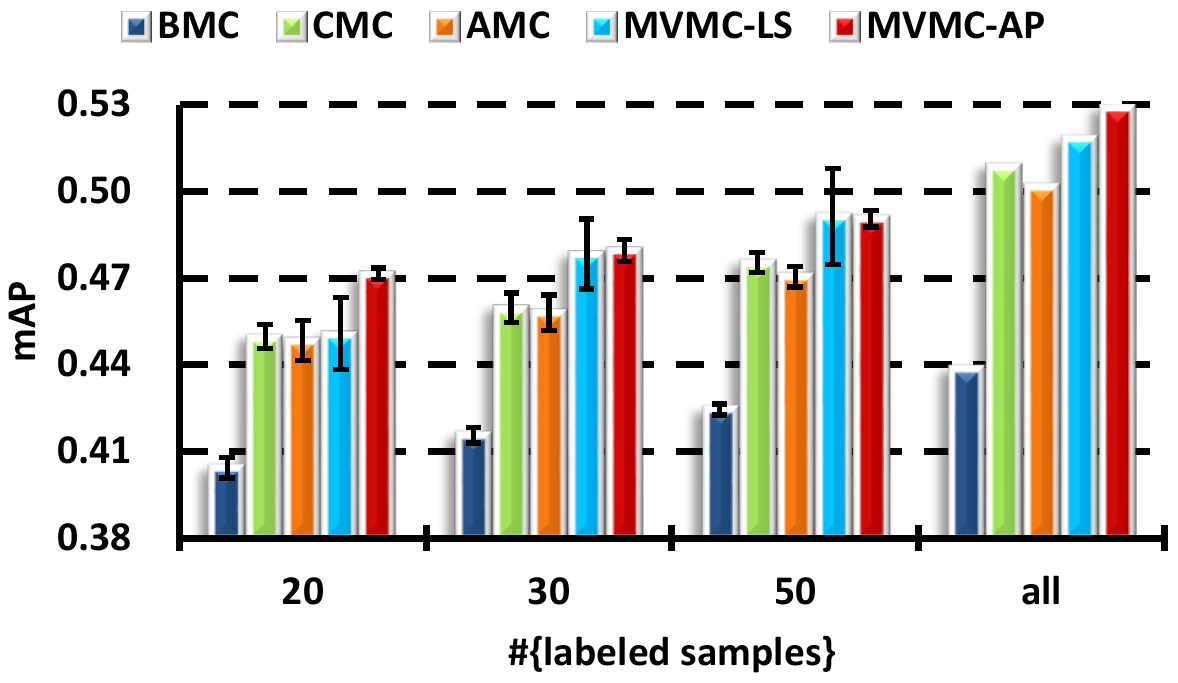}
}
\hfil
\subfigure[mAUC $\uparrow$]{\includegraphics[width=0.65\columnwidth]{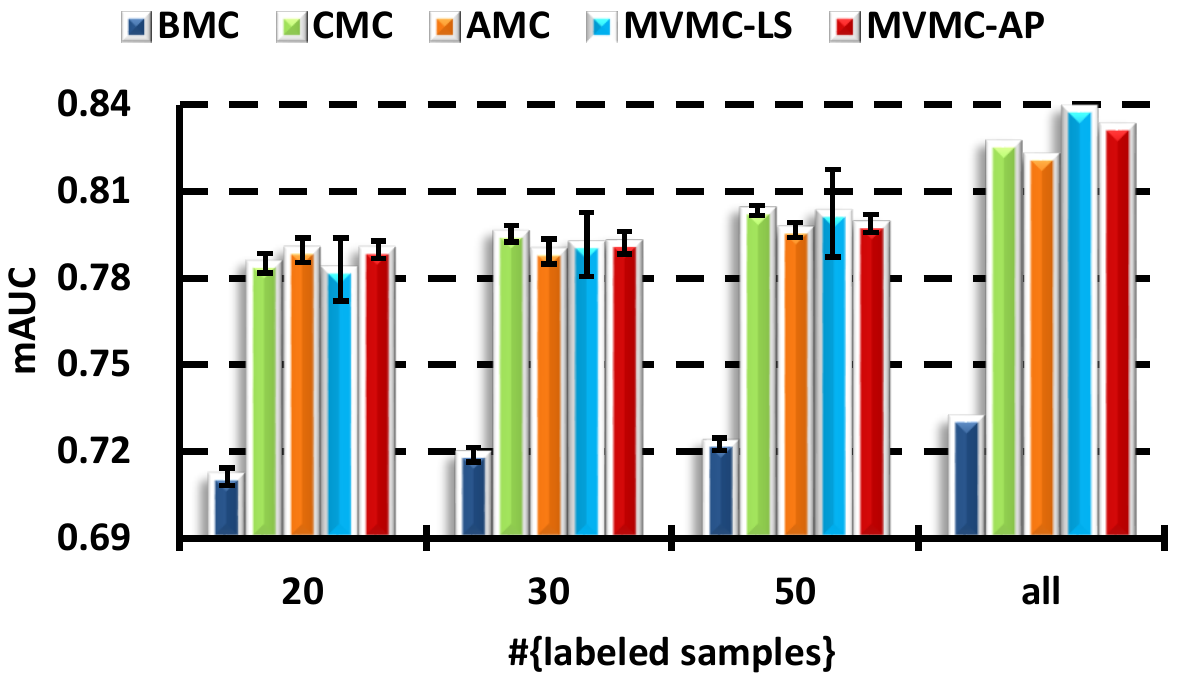}
}
\hfil
\subfigure[HL $\downarrow$]{\includegraphics[width=0.65\columnwidth]{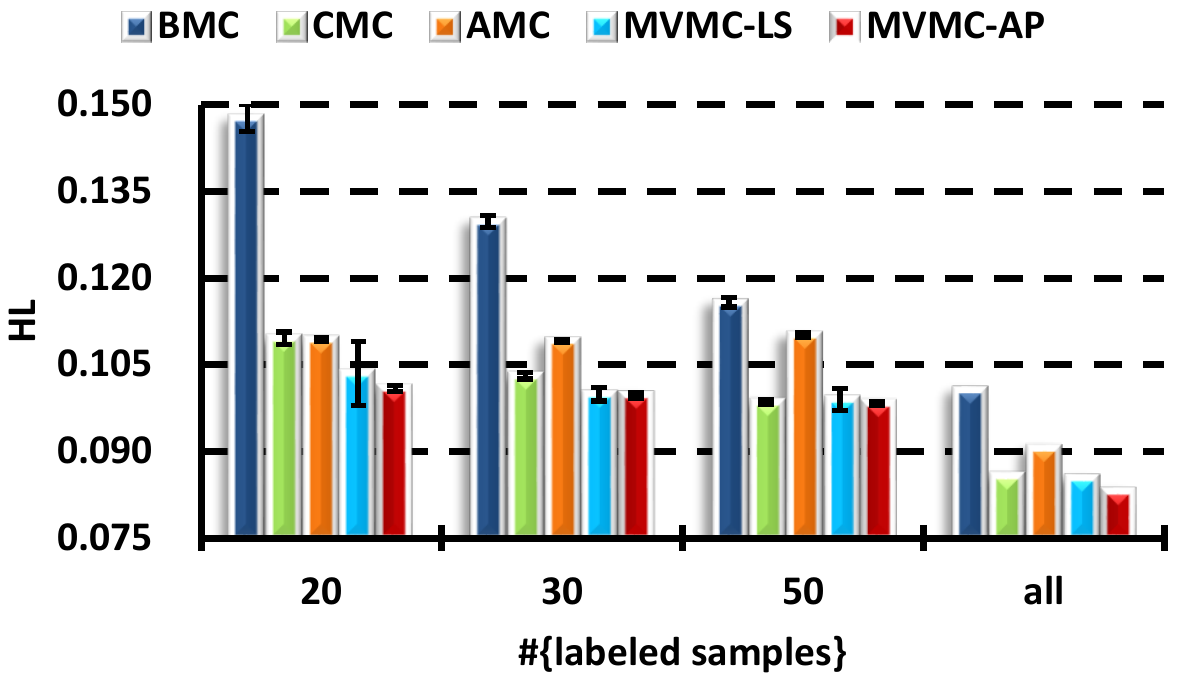}
}
\caption{A comparison of the MC-based approaches. Both the mean and standard deviation are presented. The arrow ``$\uparrow$'' indicates the larger the better, while ``$\downarrow$'' means a smaller value is better. (Top row: PASCAL VOC' 07; Bottom row: MIR Flickr.)}
\label{fig:MC_Comparison}
\end{figure*}

\subsection{Datasets, features and the evaluation criteria}
Our experiments are conducted on two popular datasets, PASCAL VOC' 07 (VOC for short)~\cite{Pascal-VOC-2007} and MIR Flickr (MIR for short)~\cite{MIR-Flickr-2008}. There are around $10,000$ natural images ($5,011$ training, $4,952$ test) from $20$ categories in the VOC dataset. The MIR dataset consists of $25,000$ images (half training, half test) and $38$ categories.

In natural image classification and web image annotation, the taxonomy of the images is very large and the category defined for a certain image may vary in different fields. Therefore, it is impossible to manually label abundant training images for every category, and it is common that limited labeled samples are given in natural image classification or annotation \cite{M-Guillaumin-et-al-CVPR-2010, ZQ-Zhao-et-al-TIP-2012}, especially for the new defined or uncommon categories. One of the advantages of the proposed transductive algorithms is the ability to handle small labeled sample size, since it can effectively utilize the information contained in the images from different views. To empirically demonstrate the effectiveness of MVMC, for both datasets, we only randomly select $n_l=\{20,30,50\}$ samples for each category in the training set as labeled, and all the others unlabeled. The labels of the unlabeled and test data are then inferred by our algorithms. Five random choices of the labeled data are used in our experiments. We also present the results of using all the labeled training samples to compare the performance of the algorithms in the fully supervised scenario. Besides, twenty percent of the test data are used for validation, which means that the parameters corresponding to the best performance on the validation set are used for unlabeled and test inference.

The features we used here are from~\cite{M-Guillaumin-et-al-CVPR-2010}, where 15 different image representations and tags are provided. Actually, the 15 representations are constructed from six kinds of visual features, which include two kinds of local features (Hue~\cite{J-Weijer-and-C-Schmid-ECCV-2006}, SIFT~\cite{D-Lowe-IJCV-2004}), a global representation (GIST~\cite{A-Oliva-and-A-Torralba-IJCV-2001}), and three global color histograms (Hsv, Lab, Rgb). In this paper, we regard each kind of visual feature as a single view. Then we have seven different views in all, with a tag view included. The dimensionality of the SIFT features we used is $1,000$, and the color features (Hsv, Lab, Rgb) is around $4,000$. The existing MC algorithms either fail to recover such a large size matrix $(25000 \times 4000)$ or the time cost is intolerable. Thus we preprocess the features by KPCA to reduce the time complexity in the MC based image classification. For all the compared methods in this paper, the result dimension after KPCA is fixed to be $50$, since we found the different algorithms perform well under this setting.

In the VOC and MIR datasets, the positive and negative samples are quite unbalanced. Thus the traditional accuracy criterion is not proper any more, and we introduce three popular criteria in multi-label classification for evaluation. They are average precision (AP)~\cite{M-Zhu-Waterloo-TR-2004}, area under ROC curve (AUC)~\cite{T-Fawcett-ML-2004} and hamming loss (HL)~\cite{R-Schapire-and-Y-Singer-ML-2000}. In this paper, AP and AUC are the ranking performance computed under each label. Usually, the mean value over all labels, i.e., mAP and mAUC are reported. HL is utilized to evaluate the label set prediction for each instance. It is not a ranking based criterion but widely used in multi-label classification \cite{ML-Zhang-and-ZH-Zhou-PR-2007}. A smaller value in HL indicates a better performance.

\subsection{A comparison with the MC-based strategies}
\label{subsec:MC_Comparison}
There are several direct strategies to make use of the different views in MC: BMC, CMC and AMC. In this section, we demonstrate that learning the output combination coefficients is superior to all of these approaches. In particular, the algorithms compared are: \\
$\bullet$ \textbf{BMC \cite{A-Goldberg-et-al-NIPS-2010, R-Cabral-et-al-NIPS-2011}:} using the best single view, i.e., one that achieves the best MC-based classification performance. The MC-based transductive multi-label classification is performed by optimizing (\ref{eq:MC_1_Formulation}), where $c_x$ is chosen to be the least squares loss for the KPCA processed features, $c_y$ is tuned with $\gamma$ in $\{1,3,30\}$. The candidate set for choosing $\lambda$ is $\{10^i \mid i = -4, \ldots ,2\}$. The parameter $\mu$ is initialized as $\mu_0 = 0.25 \sigma_1$ ($\sigma_1$: the largest singular value of $Z^0$), and decreases with a factor $0.25$ in the continuation steps until $\mu = 10^{-12}$. \\ 
$\bullet$ \textbf{CMC:} feature-level fusion. Concatenating the normalized features of all the views into a very long vector, and then performing MC-based classification. \\
$\bullet$ \textbf{AMC:} classifier-level fusion. Learning separate MC-based classifiers for different views, and then combining the outputs of all the classifiers uniformly. Before the combination, the outputs of each view are converted to the probability scores by the use of a sigmoid function $S(x) = 1 / ((1+e^{-x}))$. \\
$\bullet$ \textbf{MVMC-LS:} the proposed multi-view MC algorithm, in which the least squares loss is utilized. Outputs are also converted to probability scores. The trade-off parameter $\eta$ is tuned on the set $\{10^i \mid i = -2, \ldots ,5\}$. \\
$\bullet$ \textbf{MVMC-AP:} the proposed multi-view MC algorithm with the average precision (AP) loss.

The results for all the compared methods are presented in Fig. \ref{fig:MC_Comparison}. A self-test with different number of labeled samples is carried out to see the performance variation with respect to the labeled training size. We can see that the performance of all the presented methods improve when the number of labeled samples increase. By fusing all the views, either in the feature-level or classifier-level, can always be superior to the use of only the best single view. The concatenation (CMC) and average outputs (AMC) methods are comparable with each other. Although the proposed MVMC-LS algorithm is superior to CMC and AMC in most cases, the improvement is not significant, except for the mAP performance on the MIR dataset, while MVMC-AP consistently outperforms them in terms of all the evaluated criteria on the VOC dataset, and usually has small deviations. This demonstrates the effectiveness of the learned coefficients using the AP loss. Similar results can be seen on the MIR dataset, only the mAUC performance of the compared combination methods is comparable. In the fully supervised case (the ``all'' column), we find that CMC is superior to AMC and comparable to MVMC sometimes. This is because when large amounts of labeled data are available, the curse-of-dimension problem caused by simple concatenation is alleviated.

\subsection{Analysis of the view combination coefficients}
\label{subsec:Combination_Coefficient}

\begin{figure}
\centering
\includegraphics[width=0.8\columnwidth]{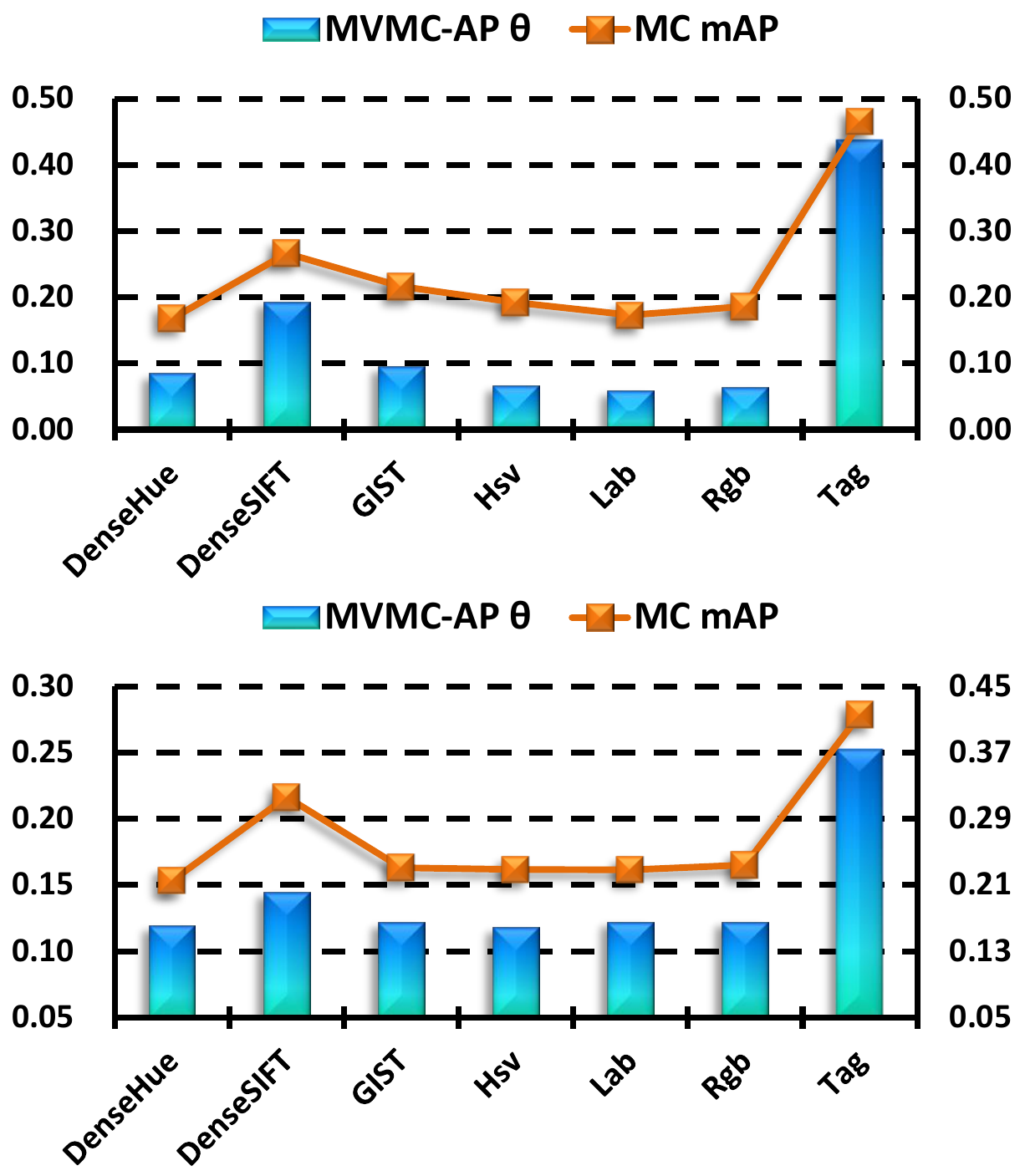}\\
\caption{The view combination coefficients learned by MVMC, together with the mAP of each MC classifier. The coefficients and the mAP scores agree in general. (Top: PASCAL VOC' 07; Bottom: MIR Flickr)}
\label{fig:Combination_Coefficient}
\end{figure}

In Fig. \ref{fig:Combination_Coefficient}, we show the combination coefficients $\theta$ learned by MVMC-AP, together with the mAP by using MC for each view. From the results, we find that the tendency of the weights is consistent with the corresponding mAP in general, i.e., the views with a higher classification performance tend to be assigned larger weights, taking the DenseSIFT (the 2nd view) and the tags (the last view) for example. The three color histogram views (Hsv, Lab, Rgb) have similar discriminative power on the VOC dataset, and thus the weights for them are almost equivalent.

\subsection{Compared with other multi-view and multi-label algorithms}
\label{subsec:Multiview_Comparison}

Our last set of experiments compare MVMC with some popular and competitive multi-view algorithms, where we learn a binary classifier for each label. Besides, extensive comparisons with the competitive and recently proposed multi-label classification algorithms are also performed. Specifically we compare MVMC-AP with the following methods: \\
$\bullet$ \textbf{HierSVM \cite{J-Kludas-et-al-AMR-2008}:} learning separate SVM classifiers for each view, and then fusing the results by using an additional SVM classifier. This is called hierarchical SVM, and the trade-off parameter $C$ for each SVM classifier is optimized over $\{10^i \mid i = -1, \ldots ,6\}$. \\
$\bullet$ \textbf{SimpleMKL \cite{A-Rakotomamonjy-et-al-JMLR-2008}:} a very popular and competitive SVM-based multiple kernel learning algorithm. Constructing a kernel for each view, and then learning a linear combination of the different kernels, as well as a classifier based on the combined kernel. The penalty factor $C$ is tuned on the set $\{10^i \mid i=-1, \ldots ,6\}$. \\
$\bullet$ \textbf{LpMKL \cite{M-Kloft-et-al-JMLR-2011}:} a recent proposed MKL algorithm, which extend MKL to $l_p$-norm with $p \geq 1$. The penalty factor $C$ is tuned on the set $\{10^i \mid i = -1, \ldots ,6\}$ and we choose the norm $p$ from the set $\{1, 8/7, 4/3, 2, 4, 8, 16, \infty\}$. \\
$\bullet$ \textbf{KLS-CCA \cite{L-Sun-et-al-TPAMI-2011}:} a least-squares formulation of the kernelized CCA for multi-label classification. The ridge parameter is chosen from the candidate set $\{10^i|i=-3,\ldots,3\}$. The different views are fused by combining the kernels (similarity matrices) with uniform weights. \\
$\bullet$ \textbf{DLP \cite{B-Wang-et-al-ICCV-2013}:} an improved label propagation algorithm that is proposed recently for transductive multi-label learning by considering the label correlation. The parameters $\alpha$ and $\lambda$ are optimized over the set $\{10^i|i=-5,\ldots,1\}$ and $\{10^i|i=-4,\ldots,2\}$ respectively. The parameter $K$ is chosen from $\{10,20,\ldots,100\}$. The different views are fused by combining the similarity matrices with uniform weights.

\begin{table*}[!t]
\setlength\tabcolsep{3pt}
\renewcommand{\arraystretch}{1.3}
\caption{A comparison of the multi-view and multi-label algorithms on the VOC dataset.}
\label{tab:Comparison_With_MultiView_Multilabel_VOC}
\centering
\begin{tabular}{c|c|c|c|c|c}
\hline
\ & \multicolumn{5}{c}{VOC \ \ mAP $\uparrow$ vs. \#\{labeled samples\}} \\
\hline \hline
Methods & 20 & 30 & 50 & all & Ranks \\
\hline
HierSVM \cite{J-Kludas-et-al-AMR-2008} & 0.459$\pm$0.032 (6) & 0.488$\pm$0.026 (5) & 0.499$\pm$0.020 (6) & 0.552 (4) & 5.25 \\
SimpleMKL \cite{A-Rakotomamonjy-et-al-JMLR-2008} & 0.490$\pm$0.006 (3) & 0.513$\pm$0.004 (3) & 0.541$\pm$0.005 (3) & 0.617 (2) & 2.75 \\
LpMKL \cite{M-Kloft-et-al-JMLR-2011} & 0.498$\pm$0.008 (2) & 0.515$\pm$0.006 (2) & 0.553$\pm$0.005 (2) & \textbf{0.618 (1)} & 1.75 \\
KLS-CCA \cite{L-Sun-et-al-TPAMI-2011} & 0.470$\pm$0.011 (5) & 0.487$\pm$0.010 (6) & 0.512$\pm$0.005 (5) & 0.538 (5) & 5.25 \\
DLP \cite{B-Wang-et-al-ICCV-2013} & 0.481$\pm$0.006 (4) & 0.499$\pm$0.004 (4) & 0.513$\pm$0.004 (4) & 0.533 (6) & 4.5 \\
MVMC-AP & \textbf{0.523$\pm$0.007 (1)} & \textbf{0.544$\pm$0.007 (1)} & \textbf{0.566$\pm$0.006 (1)} & 0.610 (3) & \textbf{1.5} \\
\hline \hline
\ & \multicolumn{5}{c}{VOC \ \ mAUC $\uparrow$ vs. \#\{labeled samples\}} \\
\hline \hline
Methods & 20 & 30 & 50 & all & Ranks \\
\hline
HierSVM \cite{J-Kludas-et-al-AMR-2008} & 0.861$\pm$0.005 (2) & 0.876$\pm$0.004 (2) & 0.886$\pm$0.005 (3) & 0.912 (3.5) & 2.63 \\
SimpleMKL \cite{A-Rakotomamonjy-et-al-JMLR-2008} & 0.848$\pm$0.005 (5) & 0.866$\pm$0.004 (4) & 0.882$\pm$0.003 (4) & 0.912 (3.5) & 4.13\\
LpMKL \cite{M-Kloft-et-al-JMLR-2011} & 0.859$\pm$0.003 (3) & 0.870$\pm$0.002 (3) & 0.889$\pm$0.003 (2) & \textbf{0.913 (1.5)} & 2.38 \\
KLS-CCA \cite{L-Sun-et-al-TPAMI-2011} & 0.831$\pm$0.012 (6) & 0.843$\pm$0.010 (6) & 0.854$\pm$0.008 (6) & 0.861 (6) & 6 \\
DLP \cite{B-Wang-et-al-ICCV-2013} & 0.852$\pm$0.002 (4) & 0.861$\pm$0.002 (5) & 0.866$\pm$0.001 (5) & 0.877 (5) & 4.75 \\
MVMC-AP & \textbf{0.872$\pm$0.001 (1)} & \textbf{0.884$\pm$0.004 (1)} & \textbf{0.894$\pm$0.002 (1)} & \textbf{0.913 (1.5)} & \textbf{1.13} \\
\hline \hline
\ & \multicolumn{5}{c}{VOC \ \ HL $\downarrow$ vs. \#\{labeled samples\}} \\
\hline \hline
Methods & 20 & 30 & 50 & all & Ranks \\
\hline
HierSVM \cite{J-Kludas-et-al-AMR-2008} & 0.059$\pm$0.003 (4) & 0.061$\pm$0.002 (3) & 0.061$\pm$0.014 (5) & 0.058 (5) & 4.25 \\
SimpleMKL \cite{A-Rakotomamonjy-et-al-JMLR-2008} & \textbf{0.052$\pm$0.002 (1.5)} & 0.051$\pm$0.001 (2) & \textbf{0.048$\pm$0.001 (1.5)} & 0.043 (2) & 1.75 \\
LpMKL \cite{M-Kloft-et-al-JMLR-2011} & 0.074$\pm$0.004 (5) & 0.077$\pm$0.005 (5) & 0.050$\pm$0.001 (3.5) & 0.050 (4) & 4.38 \\
KLS-CCA \cite{L-Sun-et-al-TPAMI-2011} & 0.056$\pm$0.001 (3) & 0.062$\pm$0.001 (4) & 0.050$\pm$0.001 (3.5) & 0.048 (3) & 3.38 \\
DLP \cite{B-Wang-et-al-ICCV-2013} & 0.229$\pm$0.000 (6) & 0.229$\pm$0.000 (6) & 0.229$\pm$0.000 (6) & 0.229 (6) & 6 \\
MVMC-AP & \textbf{0.052$\pm$0.002 (1.5)} & \textbf{0.050$\pm$0.001 (1)} & \textbf{0.048$\pm$0.001 (1.5)} & \textbf{0.042 (1)} & \textbf{1.25} \\
\hline
\end{tabular}
\begin{tabular}{m{2.0\columnwidth}}
($\uparrow$ indicates ``the larger the better''; $\downarrow$ indicates ``the smaller the better''. Mean and std. are reported. The best result is highlighted in boldface. The values in the middle and last columns are average ranks.)
\end{tabular}
\end{table*}

\begin{table*}[!t]
\setlength\tabcolsep{3pt}
\renewcommand{\arraystretch}{1.3}
\caption{A comparison of the multi-view and multi-label algorithms on the MIR dataset.}
\label{tab:Comparison_With_MultiView_Multilabel_MIR}
\centering
\begin{tabular}{c|c|c|c|c|c}
\hline
\ & \multicolumn{5}{c}{MIR \ \ mAP $\uparrow$ vs. \#\{labeled samples\}} \\
\hline \hline
Methods & 20 & 30 & 50 & all & Ranks \\
\hline
HierSVM \cite{J-Kludas-et-al-AMR-2008} & 0.422$\pm$0.011 (6) & 0.438$\pm$0.012 (6) & 0.450$\pm$0.014 (6) & 0.492 (6) & 6 \\
SimpleMKL \cite{A-Rakotomamonjy-et-al-JMLR-2008} & 0.449$\pm$0.003 (2) & 0.456$\pm$0.005 (3) & 0.470$\pm$0.003 (2) & \textbf{0.538 (1.5)} & 2.13 \\
LpMKL \cite{M-Kloft-et-al-JMLR-2011} & 0.444$\pm$0.007 (4) & 0.454$\pm$0.008 (4) & 0.468$\pm$0.009 (3) & \textbf{0.538 (1.5)} & 3.13 \\
KLS-CCA \cite{L-Sun-et-al-TPAMI-2011} & 0.436$\pm$0.007 (5) & 0.448$\pm$0.008 (5) & 0.457$\pm$0.008 (5) & 0.502 (4) & 4.75 \\
DLP \cite{B-Wang-et-al-ICCV-2013} & 0.447$\pm$0.002 (3) & 0.457$\pm$0.005 (2) & 0.464$\pm$0.007 (4) & 0.498 (5) & 3.5 \\
MVMC-AP & \textbf{0.472$\pm$0.002 (1)} & \textbf{0.480$\pm$0.004 (1)} & \textbf{0.491$\pm$0.003 (1)} & 0.529 (3) & \textbf{1.5} \\
\hline \hline
\ & \multicolumn{5}{c}{MIR \ \ mAUC $\uparrow$ vs. \#\{labeled samples\}} \\
\hline \hline
Methods & 20 & 30 & 50 & all & Ranks \\
\hline
HierSVM \cite{J-Kludas-et-al-AMR-2008} & 0.792$\pm$0.008 (5) & 0.799$\pm$0.006 (5) & 0.802$\pm$0.005 (5) & 0.836 (5) & 5 \\
SimpleMKL \cite{A-Rakotomamonjy-et-al-JMLR-2008} & 0.795$\pm$0.004 (4) & 0.801$\pm$0.005 (4) & 0.812$\pm$0.003 (4) & 0.851 (2) & 3.5 \\
LpMKL \cite{M-Kloft-et-al-JMLR-2011} & \textbf{0.809$\pm$0.003 (1.5)} & 0.810$\pm$0.003 (2.5) & 0.818$\pm$0.003 (3) & \textbf{0.854 (1)} & \textbf{2} \\
KLS-CCA \cite{L-Sun-et-al-TPAMI-2011} & 0.805$\pm$0.006 (3) & 0.810$\pm$0.006 (2.5) & \textbf{0.821$\pm$0.003 (1)} & 0.848 (3) & 2.38 \\
DLP \cite{B-Wang-et-al-ICCV-2013} & \textbf{0.809$\pm$0.002 (1.5)} & \textbf{0.815$\pm$0.003 (1)} & 0.819$\pm$0.001 (2) & 0.837 (4) & 2.13 \\
MVMC-AP & 0.790$\pm$0.003 (6) & 0.792$\pm$0.004 (6) & 0.799$\pm$0.003 (6) & 0.833 (6) & 6 \\
\hline \hline
\ & \multicolumn{5}{c}{MIR \ \ HL $\downarrow$ vs. \#\{labeled samples\}} \\
\hline \hline
Methods & 20 & 30 & 50 & all & Ranks \\
\hline
HierSVM \cite{J-Kludas-et-al-AMR-2008} & 0.105$\pm$0.009 (2) & 0.103$\pm$0.006 (2) & 0.102$\pm$0.005 (2) & 0.090 (2) & 2 \\
SimpleMKL \cite{A-Rakotomamonjy-et-al-JMLR-2008} & 0.111$\pm$0.001 (3) & 0.119$\pm$0.001 (3) & 0.114$\pm$0.001 (3) & 0.091 (3) & 3 \\
LpMKL \cite{M-Kloft-et-al-JMLR-2011} & 0.170$\pm$0.005 (5) & 0.167$\pm$0.005 (5) & 0.142$\pm$0.004 (5) & 0.119 (4) & 4.75 \\
KLS-CCA \cite{L-Sun-et-al-TPAMI-2011} & 0.123$\pm$0.000 (4) & 0.123$\pm$0.000 (4) & 0.123$\pm$0.000 (4) & 0.123 (5) & 4.25 \\
DLP \cite{B-Wang-et-al-ICCV-2013} & 0.231$\pm$0.000 (6) & 0.231$\pm$0.000 (6) & 0.231$\pm$0.000 (6) & 0.231 (6) & 6 \\
MVMC-AP & \textbf{0.101$\pm$0.001 (1)} & \textbf{0.100$\pm$0.001 (1)} & \textbf{0.098$\pm$0.001 (1)} & \textbf{0.083 (1)} & \textbf{1} \\
\hline
\end{tabular}
\begin{tabular}{m{2.0\columnwidth}}
($\uparrow$ indicates ``the larger the better''; $\downarrow$ indicates ``the smaller the better''. Mean and std. are reported. The best result is highlighted in boldface. The values in the middle and last columns are average ranks.)
\end{tabular}
\end{table*}

The performance of the compared methods on the VOC and MIR datasets are reported in Table \ref{tab:Comparison_With_MultiView_Multilabel_VOC} and \ref{tab:Comparison_With_MultiView_Multilabel_MIR}, respectively. Both the mean and standard deviation of the three criteria are presented. From the experimental results, we observe that: 1) the performance improves with an increase of the labeled samples; 2) the multi-label classification methods (KLS-CCA and DLP) are better than HierSVM, but is inferior to other multi-view learning algorithms overall; 3) DLP outperforms KLS-CCA in most cases since the unlabeled information is utilized in transduction. But when the number of labeled data is increased, the improvement deceases and sometimes KLS-CCA is better (e.g., the mAUC performance in the fully supervised case (the ``all'' column) on the MIR dataset), since the significance of the unlabeled information decreases; 4) On the VOC dataset, the mAP and HL scores of the SimpleMKL method are larger than HierSVM, while the mAUC performance of the latter is better. In general, LpMKL is superior to SimpleMKL and HierSVM. The proposed algorithm consistently outperforms the other three methods; 5) On the MIR dataset, the other methods are comparable with or superior to our algorithm in terms of mAUC, while their mAP and HL performance are poor. Under the HL criterion, SimpleMKL and HierSVM perform well on only one of the two datasets (VOC and MIR respectively), while the proposed MVMC-AP achieves the best performance consistently on both datasets. In particular, we obtain a significant $6.3\%$, $5.7\%$ and $4.9\%$ improvement in terms of mAP compared with LpMKL, when $20$, $30$ and $50$ labeled samples for each class are used, respectively; 6) In the fully supervised scenario (the ``all'' column), SimpleMKL and LpMKL are comparable to MVMC under the mAP and mAUC criteria, but the HL performance of our method is the best among all methods. Therefore, the proposed transductive classification approach is particular suitable for the small labeled sample size problem, and is comparable to the state-of-the-art methods when large amount of labeled data are available.

Besides, the average rank of the proposed alogrithm is smaller than all the other methods in terms of all the three criteria on the VOC dataset, as well as the mAP and HL criteria on the MIR dataset. According to the Friedman test~\cite{J-Demsar-JMLR-2006}, the statistics $F_F$ of mAP, mAUC and HL on the two datasets are $(16.09, 11.57)$, $(31.49, 11.79)$ and $(27.82, 137.00)$ respectively. We can see that all of them are larger than the critical value $F(5,15) = 2.27$, so we reject the null-hypothesis (the compared algorithm perform equally well). 

\section{Conclusion}
\label{sec:Conclusion}

Matrix completion (MC) has recently been used in transductive (semi-supervised) multi-label classification. It has the advantageous of being efficient, robust to noise, and being able to handle missing data. In existing algorithms, only features from a single view can be used, but there is not a single feature perfect for image classification. We therefore present a multi-view framework to fuse different kinds of features for MC-based multi-label classification. Our framework has the advantage of being able to explore the complementary properties of different views. We have designed two algorithms under the framework, MVMC-LS and MVMC-AP, which differ by the choice of different losses. The robustness of the two algorithms is analyzed.

From the experimental validation on the challenging PASCAL VOC' 07 and MIR Flickr datasets, we mainly conclude that: 1) The classifier-level fusion is better than simple feature concatenation, which is in line with \cite{C-Snoek-et-al-MM-2005, J-Kludas-et-al-AMR-2008}; 2) The learning of the combination coefficients is critical in the classifier-level fusion. Although the least squares formulation is quite efficient for learning the coefficients, the performance is usually not satisfactory. Thus more sophisticated loss should be adopted, such as the average precision loss utilized in this paper. Future works may be to extend MVMC for optimizing other criteria, such as AUC \cite{A-Herschtal-and-B-Raskutti-ICML-2004, T-Joachims-ICML-2005}, by changing the loss in MVMC.

\section{Proofs of Main Results}
\label{sec:Proofs}

In this section, we prove Theorem \ref{thm:Robustness}.
\begin{proof}
Let $\mathcal{P} \subseteq \mathbb{R}^{V \times N}, N = n_l \times m$ be the prediction space, which contains all the possible predictions. For a certain prediction $P \in \mathcal{P}$, each column is a vector $p \in \mathbb{R}^V$, which is the predictions of $V$ views for a certain sample, and a given label. The formulation (\ref{eq:Structual_SVM_Formulation}) is to find a $\theta$ to combine all the views such that the combined prediction will have the largest mean average precision over all labels.

Let $\{c_1, \ldots, c_{\mathcal{N}(\gamma/2,p,\rho)}\}$ be a $\gamma/2$ cover of $p$, where $\rho$ is a $2$-norm metric. Suppose that every prediction returned by matrix completion has the range $[-b, b]$. Then, according to Lemma \ref{lema:Banach_Subset}, we have
\begin{equation}
\notag
\mathcal{N}(\gamma/2,p,\rho) \leq (\frac{4\sqrt{V} b}{\gamma/2})^V.
\end{equation}
We can partition $\mathcal{P}$ into $(n_l!)^{m V} 2^{n_l m} \mathcal{N}(\gamma/2,p,\rho)^{n_l m}$ disjoint sets such that if $(P_1, Y_1^0)$ and $(P_2, Y_2^0)$ belong to the same set, then $Y_1^0 = Y_2^0$, rankings of different views are the same $(o_{P_1}^v = o_{P_2}^v, v = 1,\ldots,V)$ and $¦Ñ(p_1^k, p_2^k) \leq \frac{\sqrt{V} \gamma}{2}, k = 1, \ldots, N$. Here, $(n_l!)^{m V}$ is the number of all possible rankings, and $2^{n_l m}$ is the number of all possible labels (since the ground-truth label only takes two values). Note that if $P_1$ and $P_2$ have the same ranking for each view, the combination of all views will have the same ranking. Suppose that $\theta_\mathrm{ap}^*$ is the solution returned by (\ref{eq:Structual_SVM_Formulation}), then we have
\begin{equation}
|\mathrm{AP}(\theta_{\mathrm{ap}}^{*}, P_1, Y_1^0) - \mathrm{AP}(\theta_{\mathrm{ap}}^{*}, P_2, Y_2^0)| = 0.
\end{equation}
That is, the algorithm $\mathcal{A}_\mathrm{ap}$ is $(N,0)$ robust, where $N = (n_l!)^{m V} 2^{n_l m} \mathcal{N}(\gamma/2,p,\rho)^{n_l m}$.

In the next, we will prove that $\mathcal{A}_\mathrm{ls}$ is $(N,a)$ robust and $\mathcal{A}_\mathrm{svm}$ is $(N,b)$ robust, but $a,b \neq 0$. Suppose that $\theta_\mathrm{ls}^*$ is the solution returned by (\ref{eq:MVMC_LS_Formulation}), then we have
\begin{equation}
\notag
\begin{split}
& \frac{1}{2N} \sum_{k=1}^N \left( (\theta_\mathrm{ap}^*)^T p_k - y_k^0 \right)^2 + \frac{\eta}{2} \|\theta_\mathrm{ap}^*\|_2^2 \\
\leq & \frac{1}{2N} \sum_{k=1}^N \left(p_k^V - y_k^0 \right)^2 + \frac{\eta}{2} \|1\|_2^2 = \frac{\eta + (b+1)^2}{2}.
\end{split}
\end{equation}
The right side of the inequality is obtained by setting $\theta = (0, \ldots, 0, 1)$, and the inequality holds because $\theta_\mathrm{ls}^*$ is the minimizer of (\ref{eq:MVMC_LS_Formulation}). We thus further obtain $\| \theta_\mathrm{ls}^* \|_2 \leq \sqrt{((\eta+(b+1)^2)/\eta)}$. For any $(p_1,y_1^0)$ and $(p_2,y_2^0)$ in the same set of a partition, we have
\begin{equation}
\label{eq:LS_NA_Robust}
\begin{split}
& |l(\theta_\mathrm{ls}^*, (p_1,y_1^0)) - l(\theta_\mathrm{ls}^{*}, (p_2,y_2^0))| \\
= & |((\theta_\mathrm{ls}^*)^T p_1 - y_1^0)^2 - ((\theta_\mathrm{ls}^*)^T p_2 - y_2^0)^2| \\
= & |2 y_2^0 (\theta_\mathrm{ls}^*)^T p_2 - 2 y_1^0 (\theta_\mathrm{ls}^*)^T p_1 + ((\theta_\mathrm{ls}^*)^T p_1)^2 - ((\theta_\mathrm{ls}^*)^T p_2)^2| \\
\leq & |2 (\theta_\mathrm{ls}^*)^T (p_1 - p_2)| + |((\theta_\mathrm{ls}^*)^T (p_1 + p_2)) ((\theta_\mathrm{ls}^*)^T (p_1 - p_2))| \\
\leq & 2 \|\theta_\mathrm{ls}^*\|_2 \|p_1-p_2\|_2 + 2 b \|\theta_\mathrm{ls}^*\|_2 \|p_1-p_2\|_2 \\
\leq & \sqrt{V}\gamma (b+1) \sqrt{\frac{(b+1)^2+\eta}{\eta}},
\end{split}
\end{equation}
where we have utilized that $y_1^0 = y_2^0$ and $y_1^0,y_2^0 \in \{-1,1\}$. The second inequality follows from the Cauthy-Schwarz inequality. The last inequality holds because $\|p_1-p_2\|_2 \leq \sqrt{V} \gamma/2$ and $\|\theta_\mathrm{ls}^*\|_2 \leq \sqrt{(\eta+(b+1)^2)/\eta}$. Therefore, $\mathcal{A}_\mathrm{ls}$ is $(N, \gamma(b+1) \sqrt{V((b+1)^2+\eta)/\eta})$ robust. We now prove that $\mathcal{A}_\mathrm{ls}$ is not $(N,0)$ robust. Because $p$ is a vector of real values which are chosen from an infinite continuous space, there must exists a set $C'$ that contains more than three different choices of $p$ in any partition. Then for some $(p_0, y^0), (p_0 + \delta p, y^0) \in C'$, the following inequality holds
\begin{equation}
\label{eq:LS_Not_NZ_Robust}
\begin{split}
& |l(\theta_\mathrm{ls}^*, (p_0 + \delta p, y^0)) - l(\theta_\mathrm{ls}^*, (p_0, y^0))| \\
= & |2 y^0 (\theta_\mathrm{ls}^*)^T \delta p + ((\theta_\mathrm{ls}^*)^T p_0 + \delta p)^2 - ((\theta_\mathrm{ls}^*)^T p_0)^2| > 0.
\end{split}
\end{equation}
This is because there are at most two different $\delta p$ that make $2 y^0 (\theta_\mathrm{ls}^*)^T \delta p+((\theta_\mathrm{ls}^*)^T p_0 + \delta p)^2 - ((\theta_\mathrm{ls}^*)^T p_0)^2 = 0$ while the choices of $\delta p$ is more than two in the set $C'$.

According to (\ref{eq:LS_NA_Robust}) and (\ref{eq:LS_Not_NZ_Robust}), we conclude that $\mathcal{A}_\mathrm{ls}$ is $(N,a)$ robust and $a$ cannot be $0$. Similarly, we can prove that $\mathcal{A}_\mathrm{svm}$ is $(N,b)$ robust and $b \neq 0$. Therefore, the AP loss will tolerate to the small perturbation added to $p$, as long as the perturbation does not change the partition of $\mathcal{P}$, while the least squares and hinge loss will not. This completes the proof.
\end{proof}


%




\ifCLASSOPTIONcaptionsoff
  \newpage
\fi



\scriptsize
\bibliographystyle{IEEEtran}
\bibliography{./TIP-12151-2014}

\begin{thebibliography}{10}
\providecommand{\url}[1]{#1}
\csname url@samestyle\endcsname
\providecommand{\newblock}{\relax}
\providecommand{\bibinfo}[2]{#2}
\providecommand{\BIBentrySTDinterwordspacing}{\spaceskip=0pt\relax}
\providecommand{\BIBentryALTinterwordstretchfactor}{4}
\providecommand{\BIBentryALTinterwordspacing}{\spaceskip=\fontdimen2\font plus
\BIBentryALTinterwordstretchfactor\fontdimen3\font minus
  \fontdimen4\font\relax}
\providecommand{\BIBforeignlanguage}[2]{{%
\expandafter\ifx\csname l@#1\endcsname\relax
\typeout{** WARNING: IEEEtran.bst: No hyphenation pattern has been}%
\typeout{** loaded for the language `#1'. Using the pattern for}%
\typeout{** the default language instead.}%
\else
\language=\csname l@#1\endcsname
\fi
#2}}
\providecommand{\BIBdecl}{\relax}
\BIBdecl

\bibitem{M-Boutell-et-al-PR-2004}
M.~Boutell, J.~Luo, X.~Shen, and C.~Brown, ``Learning multi-label scene
  classification,'' \emph{Pattern Recognition}, vol.~37, no.~9, pp. 1757--1771,
  2004.

\bibitem{G-Tsoumakas-and-I-Katakis-IJDWM-2007}
G.~Tsoumakas and I.~Katakis, ``Multi-label classification: An overview,''
  \emph{International Journal of Data Warehousing and Mining}, vol.~3, no.~3,
  pp. 1--13, 2007.

\bibitem{B-Hariharan-et-al-ICML-2010}
B.~Hariharan, L.~Zelnik-Manor, S.~Vishwanathan, and M.~Varma, ``Large scale
  max-margin multi-label classification with priors,'' in \emph{International
  Conference on Machine Learning}, 2010, pp. 423--430.

\bibitem{L-Sun-et-al-TPAMI-2011}
L.~Sun, S.~Ji, and J.~Ye, ``Canonical correlation analysis for multilabel
  classification: a least-squares formulation, extensions, and analysis,''
  \emph{IEEE Transactions on Pattern Analysis and Machine Intelligence},
  vol.~33, no.~1, pp. 194--200, 2011.

\bibitem{B-Wang-et-al-ICCV-2013}
B.~Wang, Z.~Tu, and J.~K. Tsotsos, ``Dynamic label propagation for
  semi-supervised multi-class multi-label classification,'' in \emph{IEEE
  International Conference on Computer Vision}, 2013, pp. 425--432.

\bibitem{G-Sundaramoorthi-and-BW-Hong-CVPR-2014}
G.~Sundaramoorthi and B.-W. Hong, ``Fast label: Easy and efficient solution of
  joint multi-label and estimation problems,'' in \emph{IEEE Conference on
  Computer Vision and Pattern Recognition}, 2014, pp. 3126--3133.

\bibitem{A-Goldberg-et-al-NIPS-2010}
A.~Goldberg, X.~Zhu, B.~Recht, J.~Xu, and R.~Nowak, ``Transduction with matrix
  completion: Three birds with one stone,'' in \emph{Advances in Neural
  Information Processing Systems}, 2010, pp. 757--765.

\bibitem{R-Cabral-et-al-NIPS-2011}
R.~Cabral, F.~De~la Torre, J.~Costeira, and A.~Bernardino, ``Matrix completion
  for multi-label image classification,'' in \emph{Advances in Neural
  Information Processing Systems}, 2011, pp. 190--198.

\bibitem{J-Cai-et-al-SIAM-2010}
J.~Cai, E.~Cand{\`e}s, and Z.~Shen, ``A singular value thresholding algorithm
  for matrix completion,'' \emph{SIAM Journal on Optimization}, vol.~20, no.~4,
  pp. 1956--1982, 2010.

\bibitem{Pascal-VOC-2007}
M.~Everingham, L.~Van~Gool, C.~K.~I. Williams, J.~Winn, and A.~Zisserman, ``The
  {PASCAL} {V}isual {O}bject {C}lasses {C}hallenge 2007 {(VOC2007)}
  {R}esults,'' 2007.

\bibitem{MIR-Flickr-2008}
M.~J. Huiskes and M.~S. Lew, ``The {MIR} flickr retrieval evaluation,'' in
  \emph{International conference on Multimedia Information Retrieval}, 2008,
  pp. 39--43.

\bibitem{A-Rakotomamonjy-et-al-JMLR-2008}
A.~Rakotomamonjy, F.~Bach, S.~Canu, and Y.~Grandvalet, ``Simple{MKL},''
  \emph{Journal of Machine Learning Research}, vol.~9, pp. 2491--2521, 2008.

\bibitem{M-Kloft-et-al-JMLR-2011}
M.~Kloft, U.~Brefeld, S.~Sonnenburg, and A.~Zien, ``Lp-norm multiple kernel
  learning,'' \emph{Journal of Machine Learning Research}, vol.~12, pp.
  953--997, 2011.

\bibitem{J-Kludas-et-al-AMR-2008}
J.~Kludas, E.~Bruno, and S.~Marchand-Maillet, ``Information fusion in
  multimedia information retrieval,'' \emph{Adaptive Multimedial Retrieval:
  Retrieval, User, and Semantics}, pp. 147--159, 2008.

\bibitem{S-Ono-et-al-TIP-2014}
S.~Ono, T.~Miyata, and I.~Yamada, ``Cartoon-texture image decomposition using
  blockwise low-rank texture characterization,'' \emph{IEEE Transactions on
  Image Processing}, vol.~23, no.~3, pp. 1128--1142, 2014.

\bibitem{H-Arguello-and-GR-Arce-TIP-2013}
H.~Arguello and G.~R. Arce, ``Rank minimization code aperture design for
  spectrally selective compressive imaging,'' \emph{IEEE Transactions on Image
  Processing}, vol.~22, no.~3, pp. 941--954, 2013.

\bibitem{E-Candes-and-B-Recht-FCM-2009}
E.~Cand{\`e}s and B.~Recht, ``Exact matrix completion via convex
  optimization,'' \emph{Foundations of Computational Mathematics}, vol.~9,
  no.~6, pp. 717--772, 2009.

\bibitem{M-Fazel-Thesis-2002}
M.~Fazel, ``Matrix rank minimization with applications,'' Ph.D. dissertation,
  Stanford University, 2002.

\bibitem{SQ-Ma-et-al-MP-2011}
S.~Ma, D.~Goldfarb, and L.~Chen, ``Fixed point and bregman iterative methods
  for matrix rank minimization,'' \emph{Mathematical Programming}, vol. 128,
  no.~1, pp. 321--353, 2011.

\bibitem{R-Keshavan-et-al-TIT-2010}
R.~H. Keshavan, A.~Montanari, and S.~Oh, ``Matrix completion from a few
  entries,'' \emph{IEEE Transactions on Information Theory}, vol.~56, no.~6,
  pp. 2980--2998, 2010.

\bibitem{ZC-Lin-et-al-NIPS-2011}
Z.~Lin, R.~Liu, and Z.~Su, ``Linearized alternating direction method with
  adaptive penalty for low-rank representation,'' in \emph{Advances in Neural
  Information Processing Systems}, 2011, pp. 612--620.

\bibitem{ZC-Lin-et-al-TR-UILU-2009}
Z.~Lin, A.~Ganesh, J.~Wright, L.~Wu, M.~Chen, and Y.~Ma, ``Fast convex
  optimization algorithms for exact recovery of a corrupted low-rank matrix,''
  University of Illinois, Urbana, Tech. Rep. UILU-ENG-09-2214, 2009.

\bibitem{A-Zien-and-C-Ong-ICML-2007}
A.~Zien and C.~S. Ong, ``Multiclass multiple kernel learning,'' in
  \emph{International Conference on Machine Learning}, 2007, pp. 1191--1198.

\bibitem{S-Bickel-and-T-Scheffer-ICDM-2004}
S.~Bickel and T.~Scheffer, ``Multi-view clustering,'' in \emph{International
  Conference on Data Mining}, 2004, pp. 19--26.

\bibitem{A-Rodriguez-et-al-TIP-2013}
A.~Rodriguez, V.~N. Boddeti, B.~V. Kumar, and A.~Mahalanobis, ``Maximum margin
  correlation filter: A new approach for localization and classification,''
  \emph{IEEE Transactions on Image Processing}, vol.~22, no.~2, pp. 631--643,
  2013.

\bibitem{J-Lindblad-and-N-Sladoje-TIP-2014}
J.~Lindblad and N.~Sladoje, ``Linear time distances between fuzzy sets with
  applications to pattern matching and classification,'' \emph{IEEE
  Transactions on Image Processing}, vol.~23, no.~1, pp. 126--136, 2014.

\bibitem{R-Ptucha-and-AE-Savakis-TIP-2014}
R.~Ptucha and A.~E. Savakis, ``{LGE-KSVD}: robust sparse representation
  classification,'' \emph{IEEE Transactions on Image Processing}, vol.~23,
  no.~4, pp. 1737--1750, 2014.

\bibitem{G-Lanckriet-et-al-ICML-2002}
G.~Lanckriet, N.~Cristianini, P.~Bartlett, L.~Ghaoui, and M.~Jordan, ``Learning
  the kernel matrix with semidefinite programming,'' in \emph{International
  Conference on Machine Learning}, 2002, pp. 323--330.

\bibitem{G-Lanckriet-et-al-JMLR-2004}
------, ``Learning the kernel matrix with semidefinite programming,''
  \emph{Journal of Machine Learning Research}, vol.~5, pp. 27--72, 2004.

\bibitem{Y-Lin-et-al-NIPS-2008}
Y.~Y. Lin, T.~L. Liu, and C.~S. Fuh, ``Dimensionality reduction for data in
  multiple feature representations,'' in \emph{Advances in Neural Information
  Processing Systems}, 2008, pp. 961--968.

\bibitem{B-McFee-and-G-Lanckriet-JMLR-2011}
B.~McFee and G.~Lanckriet, ``Learning multi-modal similarity,'' \emph{Journal
  of Machine Learning Research}, vol.~12, pp. 491--523, 2011.

\bibitem{D-Hardoon-et-al-NCn-2004}
D.~R. Hardoon, S.~Szedmak, and J.~Shawe-Taylor, ``Canonical correlation
  analysis: An overview with application to learning methods,'' \emph{Neural
  Computation}, vol.~16, no.~12, pp. 2639--2664, 2004.

\bibitem{J-Farquhar-et-al-NIPS-2005}
J.~Farquhar, D.~Hardoon, H.~Meng, J.~S. Shawe-taylor, and S.~Szedmak, ``Two
  view learning: {SVM-2K}, theory and practice,'' in \emph{Advances in Neural
  Information Processing Systems}, 2005, pp. 355--362.

\bibitem{M-White-et-al-NIPS-2012}
M.~White, X.~Zhang, D.~Schuurmans, and Y.-l. Yu, ``Convex multi-view subspace
  learning,'' in \emph{Advances in Neural Information Processing Systems},
  2012, pp. 1682--1690.

\bibitem{C-Snoek-et-al-MM-2005}
C.~G. Snoek, M.~Worring, and A.~W. Smeulders, ``Early versus late fusion in
  semantic video analysis,'' in \emph{ACM International Conference on
  Multimedia}, 2005, pp. 399--402.

\bibitem{G-Fumera-and-F-Roli-TPAMI-2005}
G.~Fumera and F.~Roli, ``A theoretical and experimental analysis of linear
  combiners for multiple classifier systems,'' \emph{IEEE Transactions on
  Pattern Analysis and Machine Intelligence}, vol.~27, no.~6, pp. 942--956,
  2005.

\bibitem{M-Wozniak-and-K-Jackowski-HAIS-2009}
M.~Wozniak and K.~Jackowski, ``Some remarks on chosen methods of classifier
  fusion based on weighted voting,'' in \emph{Hybrid Artificial Intelligence
  Systems}, 2009, pp. 541--548.

\bibitem{A-Blum-and-T-Mitchell-COLT-1998}
A.~Blum and T.~Mitchell, ``Combining labeled and unlabeled data with
  co-training,'' in \emph{Annual conference on Computational Learning Theory},
  1998, pp. 92--100.

\bibitem{B-Krishnapuram-et-al-NIPS-2004}
B.~Krishnapuram, D.~Williams, Y.~Xue, L.~Carin, M.~Figueiredo, and A.~J.
  Hartemink, ``On semi-supervised classification,'' in \emph{Advances in Neural
  Information Processing Systems}, 2004, pp. 721--728.

\bibitem{C-Christoudias-et-al-CVPR-2009}
C.~M. Christoudias, R.~Urtasun, A.~Kapoorz, and T.~Darrell, ``Co-training with
  noisy perceptual observations,'' in \emph{IEEE conference on Computer Vision
  and Pattern Recognition}, 2009, pp. 2844--2851.

\bibitem{K-Nigam-and-R-Ghani-CIKM-2000}
K.~Nigam and R.~Ghani, ``Analyzing the effectiveness and applicability of
  co-training,'' in \emph{International Conference on Information and Knowledge
  Management}, 2000, pp. 86--93.

\bibitem{U-Brefeld-and-T-Scheffer-ICML-2004}
U.~Brefeld and T.~Scheffer, ``{Co-EM} support vector learning,'' in
  \emph{International Conference on Machine Learning}, 2004, pp. 121--128.

\bibitem{V-Sindhwani-et-al-ICMLw-2005}
V.~Sindhwani, P.~Niyogi, and M.~Belkin, ``A co-regularization approach to
  semi-supervised learning with multiple views,'' in \emph{ICML Workshop on
  Learning with Multiple Views}, 2005, pp. 74--79.

\bibitem{M-Belkin-et-al-JMLR-2006}
M.~Belkin, P.~Niyogi, and V.~Sindhwani, ``Manifold regularization: A geometric
  framework for learning from labeled and unlabeled examples,'' \emph{Journal
  of Machine Learning Research}, vol.~7, pp. 2399--2434, 2006.

\bibitem{A-Klausner-et-al-ICDSC-2007}
A.~Klausner, A.~Tengg, and B.~Rinner, ``Vehicle classification on multi-sensor
  smart cameras using feature-and decision-fusion,'' in \emph{ACM/IEEE
  International Conference on Distributed Smart Cameras}, 2007, pp. 67--74.

\bibitem{D-Lowe-IJCV-2004}
D.~Lowe, ``Distinctive image features from scale-invariant keypoints,''
  \emph{International Journal of Computer Vision}, vol.~60, no.~2, pp. 91--110,
  2004.

\bibitem{A-Oliva-and-A-Torralba-IJCV-2001}
A.~Oliva and A.~Torralba, ``Modeling the shape of the scene: A holistic
  representation of the spatial envelope,'' \emph{International Journal of
  Computer Vision}, vol.~42, no.~3, pp. 145--175, 2001.

\bibitem{Pascal-VOC}
\url{http://pascallin.ecs.soton.ac.uk/challenges/VOC/}.

\bibitem{M-Guillaumin-et-al-CVPR-2010}
M.~Guillaumin, J.~Verbeek, and C.~Schmid, ``Multimodal semi-supervised learning
  for image classification,'' in \emph{IEEE conference on Computer Vision and
  Pattern Recognition}, 2010, pp. 902--909.

\bibitem{A-Christmann-and-I-Steinwart-JSTOR-2007}
A.~Christmann and I.~Steinwart, ``Consistency and robustness of kernel-based
  regression in convex risk minimization,'' \emph{Bernoulli}, pp. 799--819,
  2007.

\bibitem{YC-Wu-and-YF-Liu-JASA-2007}
Y.~Wu and Y.~Liu, ``Robust truncated hinge loss support vector machines,''
  \emph{Journal of the American Statistical Association}, vol. 102, no. 479,
  pp. 974--983, 2007.

\bibitem{C-Manning-et-al-Cambridge-Book-2008}
C.~D. Manning, P.~Raghavan, and H.~Sch{\"u}tze, \emph{Introduction to
  information retrieval}.\hskip 1em plus 0.5em minus 0.4em\relax Cambridge
  University Press, 2008.

\bibitem{Y-Yue-et-al-SIGIR-2007}
Y.~Yue, T.~Finley, F.~Radlinski, and T.~Joachims, ``A support vector method for
  optimizing average precision,'' in \emph{ACM SIGIR conference on research and
  development in information retrieval}, 2007, pp. 271--278.

\bibitem{I-Tsochantaridis-et-al-JMLR-2005}
I.~Tsochantaridis, T.~Joachims, T.~Hofmann, and Y.~Altun, ``Large margin
  methods for structured and interdependent output variables,'' \emph{Journal
  of Machine Learning Research}, pp. 1453--1484, 2005.

\bibitem{T-Joachims-et-al-MLJ-2009}
T.~Joachims, T.~Finley, and C.-N.~J. Yu, ``Cutting-plane training of structural
  svms,'' \emph{Machine Learning}, vol.~77, no.~1, pp. 27--59, 2009.

\bibitem{H-Xu-and-S-Mannor-MLJ-2012}
H.~Xu and S.~Mannor, ``Robustness and generalization,'' \emph{Machine
  Learning}, vol.~86, no.~3, pp. 391--423, 2012.

\bibitem{K-Aas-and-L-Eikvil-TR-Norwegian-1999}
K.~Aas and L.~Eikvil, ``Text categorization: a survey,'' Norwegian Computing
  Center, Tech. Rep., 1999.

\bibitem{ML-Zhang-and-ZH-Zhou-PR-2007}
M.-L. Zhang and Z.-H. Zhou, ``{ML-KNN}: A lazy learning approach to multi-label
  learning,'' \emph{Pattern Recognition}, vol.~40, no.~7, pp. 2038--2048, 2007.

\bibitem{HQ-Minh-and-V-Sindhwani-ICML-2011}
H.~Q. Minh and V.~Sindhwani, ``Vector-valued manifold regularization,'' in
  \emph{International Conference on Machine Learning}, 2011, pp. 57--64.

\bibitem{T-Poggio-and-CR-Shelton-AMS-2002}
T.~Poggio and C.~Shelton, ``On the mathematical foundations of learning,''
  \emph{American Mathematical Society}, vol.~39, no.~1, pp. 1--49, 2002.

\bibitem{A-Maurer-and-M-Pontil-TIT-2010}
A.~Maurer and M.~Pontil, ``K-dimensional coding schemes in hilbert spaces,''
  \emph{IEEE Transactions on Information Theory}, vol.~56, no.~11, pp.
  5839--5846, 2010.

\bibitem{ZQ-Zhao-et-al-TIP-2012}
Z.-Q. Zhao, H.~Glotin, Z.~Xie, J.~Gao, and X.~Wu, ``Cooperative sparse
  representation in two opposite directions for semi-supervised image
  annotation,'' \emph{IEEE Transactions on Image Processing}, vol.~21, no.~9,
  pp. 4218--4231, 2012.

\bibitem{J-Weijer-and-C-Schmid-ECCV-2006}
J.~Van De~Weijer and C.~Schmid, ``Coloring local feature extraction,'' in
  \emph{European Conference on Computer Vision}, 2006, pp. 334--348.

\bibitem{M-Zhu-Waterloo-TR-2004}
M.~Zhu, ``Recall, precision and average precision,'' Dept. Electr. Eng., Univ.
  Waterloo, Tech. Rep. Working Paper 2004-09, 2004.

\bibitem{T-Fawcett-ML-2004}
T.~Fawcett, ``{ROC} graphs: Notes and practical considerations for
  researchers,'' \emph{Machine Learning}, vol.~31, pp. 1--38, 2004.

\bibitem{R-Schapire-and-Y-Singer-ML-2000}
R.~Schapire and Y.~Singer, ``Boostexter: A boosting-based system for text
  categorization,'' \emph{Machine Learning}, vol.~39, no.~2, pp. 135--168,
  2000.

\bibitem{J-Demsar-JMLR-2006}
J.~Dem{\v{s}}ar, ``Statistical comparisons of classifiers over multiple data
  sets,'' \emph{Journal of Machine Learning Research}, vol.~7, pp. 1--30, 2006.

\bibitem{A-Herschtal-and-B-Raskutti-ICML-2004}
A.~Herschtal and B.~Raskutti, ``Optimising area under the roc curve using
  gradient descent,'' in \emph{International Conference on Machine Learning},
  2004, pp. 49--56.

\bibitem{T-Joachims-ICML-2005}
T.~Joachims, ``A support vector method for multivariate performance measures,''
  in \emph{International Conference on Machine Learning}, 2005, pp. 377--384.

\end{thebibliography}
%
%
%

%




\end{document}